\documentclass[dvipsnames]{article} % For LaTeX2e
\usepackage{iclr2024_conference,times}
\iclrfinalcopy
\usepackage{fullpage} % this cancels the header!!!!!!!!

\usepackage{hyperref}
\usepackage{url}
\usepackage{enumerate}
\usepackage{amscd, amsmath, amsbsy,amsfonts, amssymb, amsthm, bm, bbm,mathrsfs}
\usepackage{graphicx}
\usepackage{subcaption}
\usepackage[capitalize,noabbrev]{cleveref}
\usepackage{tikz}
\usepackage{transparent}

\newtheorem{othertheorem}{othertheorem}[section]

\theoremstyle{definition}
\newtheorem{definition}[othertheorem]{Definition}
\newtheorem{remark}[othertheorem]{Remark}

\theoremstyle{definition}

\usepackage{wrapfig}
\usepackage{multirow}
\usepackage{multicol}
\graphicspath{{figures/}}
\usepackage{varwidth}

\usepackage{soul}
\usepackage{tikz-cd}

\newcommand{\x}{\bm{x}}
\renewcommand{\a}{\bm{a}}
\newcommand{\s}{\bm{s}}

\newcommand{\g}{\bm{g}}
\newcommand{\y}{\bm{y}}
\newcommand{\z}{\bm{z}}

\renewcommand{\b}{\mathbf{b}}
\newcommand{\I}{\mathbf{I}}

\newcommand{\R}{\mathbb{R}}

\newcommand{\W}{\mathbf{W}}
\renewcommand{\s}{\bm{s}}

\newcommand{\G}{\mathbf{G}}

\usepackage{pifont}% http://ctan.org/pkg/pifont
\usepackage[noend]{algpseudocode}

\title{Zero redundancy distributed learning with differential privacy}
\author{Zhiqi Bu, Justin Chiu, Ruixuan Liu, Sheng Zha, George Karypis}
\date{}

\author{Zhiqi Bu\thanks{\texttt{zhiqibu@amazon.com}} \& Ruixuan Liu \& Justin Chiu \& Sheng Zha \& George Karypis\\
Amazon Web Services AI
}

\begin{document}
\maketitle

% \vspace{-0.7cm}
\begin{abstract}
Deep learning using large models have achieved great success in a wide range of domains. However, training these models on billions of parameters is very challenging in terms of the training speed, memory cost, and communication efficiency, especially under the privacy-preserving regime with differential privacy (DP). On the one hand, DP optimization has comparable efficiency to the standard non-private optimization on a single GPU, but on multiple GPUs, existing DP distributed learning (such as pipeline parallel) has suffered from significantly worse efficiency. On the other hand, the Zero Redundancy Optimizer (ZeRO) is a state-of-the-art solution to the standard distributed learning, exhibiting excellent training efficiency on large models, but to work compatibly with DP is technically complicated. In this work, we develop a new systematic solution, DP-ZeRO, (I) to scale up the trainable DP model size, e.g. to GPT-100B, (II) to obtain the same computation and communication efficiency as the standard ZeRO, and (III) to enable mixed-precision DP training. Our DP-ZeRO, like the standard ZeRO, has the potential to train models with arbitrary size and is evaluated on the world's largest DP models in terms of the number of trainable parameters.
% \footnote{Code at \url{https://github.com/awslabs/fast-differential-privacy}.}
\end{abstract}

\section{Introduction}
\label{sec:intro}
Recent advances in differentially private (DP) deep learning have witnessed the power of large pre-trained models, achieving comparable accuracy to state-of-the-art (SOTA) non-private models across computer vision \cite{de2022unlocking,bu2022scalable,mehta2022large,xie2018differentially}, natural language processing \cite{yu2021differentially,li2021large,bu2022automatic}, and many other tasks. Similar to their non-DP counter-parts, it has been observed that larger DP models tend to have better performance. For example, the DP accuracy increases from 83\% using RoBERTa-base (123M parameters) to 86\% using RoBERTa-large (354M parameters) on GLUE datasets \cite{li2021large,bu2022automatic,yu2021differentially}; the DP BLEU score increases from 61 using GPT2-small (124M parameters) to 64 using GPT2-large (800M parameters) on E2E dataset \cite{li2021large,bu2022automatic}; a similar trend is also observed using ViT (Base/Large/Huge) up to 600M parameters to achieve state-of-the-art DP accuracy on ImageNet, around 81\% at $\epsilon=8$ \cite{mehta2022large}. 

Driven by this success and the surge of computational power, it is high time to enable DP deep learning at the same scale of the standard non-DP one, e.g., GPT3-175B \citep{brown2020language} and LLaMA-63B \citep{touvron2022llama,touvron2023llama} with billions of trainable parameters. Specifically, such a DP training system must have high time and memory efficiency, low communication cost, and the compatibility with general neural network architectures. 

For small to moderately large models (e.g. with less than a billion parameters) that fit within the memory of a single GPU, a range of DP algorithms are feasible, producing the same result at different efficiency. Examples include TensorFlow-privacy \cite{subramani2021enabling}, Opacus \cite{opacus,bu2022dpbitfit}, ghost clipping (GhostClip) \cite{goodfellow2015efficient,li2021large,bu2022scalable}, and Book-Keeping (BK) \cite{bu2022differentially}, among which the BK algorithm has allowed DP optimization to be almost as efficient as the standard one. To be specific, the time/space complexity of BK algorithm is $1.08\times/1.05\times$ of the standard optimization on ViT-Large (300M parameters, 147 layers) and $1.03\times/1.01\times$ on GPT2-large (800M parameters, 220 layers). 

To enable the DP distributed learning of these not-too-large models, one can directly use DDP (distributed data parallelism) \citep{li13pytorch}, where each mini-batch of data is partitioned to smaller micro-batches and each GPU computes one micro-batch with a full copy of the DP model. A line of researches \citep{opacus,de2022unlocking,kurakin2022toward} have reported that DDP with DP usually either incurs huge memory cost due to caching the per-sample gradients, or suffers from $2-9\times$ slower training speed than non-DP optimization \cite{de2022unlocking,bu2021fast}. While the efficiency issues can be addressed through a better DP algorithm, such as BK, the feasibility issue remains insurmountable because DDP cannot train models that exceed the capacity of one GPU. Notably, the efficiency of BK algorithm is enhanced by two key techniques: \textit{mixed ghost norm} (computing per-sample gradient norms almost for free) and \textit{book-keeping trick} (only using one round of full back-propagation, not two rounds as in \cite{li2021large,bu2022scalable}), which are detailed in \Cref{app:BK} and will also be leveraged in our DP-ZeRO solution.

As the model size further increases beyond a reasonable bound for one GPU (e.g. 32GB memory, which roughly translates to 2B model training with Adam), the model must be partitioned in addition to the data, e.g. using pipeline parallelism and model parallelism, so that each GPU only holds a partial shard of the model (see \Cref{fig:grad partition}). In \cite{anonymous2023exploring}, DP is combined with pipeline parallelism to fine-tune about 0.1\% of GPT3-175B. Yet, the pipeline parallelism can be inefficient due to a non-DP-related issue, known as the pipeline bubble, where GPUs are idle while waiting for data to process.
 
\begin{table}[!htb]
    \centering
\vspace{-0.3cm}
\caption{Summary of DP distributed learning.}
\vspace{-0.2cm}
\resizebox{\linewidth}{!}{
\begin{tabular}{c|c|c|c|c|c}
% \multirow{2}{*}{\shortstack{ Distributed\\solution}}&\multirow{2}{*}{Parallelism}&\multirow{2}{*}{\shortstack{Model\\ sharding}}&\multicolumn{2}{c|}{Related work}&\multirow{2}{*}{DP solution}&\multirow{2}{*}{Bottleneck}&\multirow{2}{*}{Remark}\\
% &&&standard&DP&&\\\hline
% DDP &Data&No&\cite{li13pytorch}&\cite{opacus}&Opacus&memory&cannot fit large model; memory inefficient (due to per-sample gradients)\\
% DDP&Data&No&\cite{bradbury2018jax,frostig2018compiling}&\cite{de2022unlocking,kurakin2022toward}&JAX\cite{subramani2021enabling}&memory \& speed& cannot fit large model; time inefficient (e.g. $2-9\times$ slowdown)\\
% GPipe&Pipeline&Yes&\cite{huang2019gpipe}&\cite{anonymous2023exploring}&GhostClip\cite{li2021large}&speed& pipeline parallelism has bubble that wastes GPU time\\
% ZeRO&Data(\&Model)&Yes&\cite{rajbhandari2020zero,rasley2020deepspeed}&\textbf{Ours}&BK\cite{bu2022differentially}&---&time \& memory \& communication efficient\\

\multirow{2}{*}{\shortstack{ Distributed\\solution}}&\multirow{2}{*}{Parallelism}&\multirow{2}{*}{\shortstack{Model\\ sharding}}&\multirow{2}{*}{\shortstack{Standard\\ version}}&\multirow{2}{*}{\shortstack{DP\\ version}}&\multirow{2}{*}{Remark}\\
&&&&&\\\hline
DDP &Data&No&\cite{li13pytorch}&\cite{opacus}&unable to fit large model and DP is memory costly  \\
DDP&Data&No&\cite{frostig2018compiling}&\cite{de2022unlocking}&unable to fit large model and DP is slow\\
GPipe&Pipeline&Yes&\cite{huang2019gpipe}&\cite{anonymous2023exploring}&pipeline bubble wastes GPU time\\
ZeRO&Data(\&Model)&Yes&\cite{rajbhandari2020zero}&\textbf{Ours}&speed \& memory \& communication efficient\\
\end{tabular}
}
\label{tab:dp distrib}
\end{table}

Generally speaking, more advanced distributed methods such as Zero Redundancy Optimizer \cite{rajbhandari2020zero} (\textbf{ZeRO}) and mixed-precision training have not be paired with DP due to the lack of algorithmic advances. In this work, we develop DP-ZeRO, equipping state-of-the-art distributed learning solution with DP (see comparison in \Cref{tab:dp distrib}), without altering the mathematics of DP optimization. We summarize our contributions as follows.

\begin{enumerate}
    \item We propose the zero redundancy distributed learning with differential privacy (DP-ZeRO), demonstrating the same level of \textbf{communication efficiency}, \textbf{computation efficiency (speed and memory)}, and \textbf{scalability} (e.g. to GPT3 level and hundreds of GPUs) as the standard ZeRO.
    % , which is not possible on a single GPU.
    \item We enable the mixed-precision training with DP by addressing the issue of loss scaling. This advance allows us to reduce the memory cost by roughly 50\% and allow significantly faster communication that was previously not enjoyed by DP distributed learning.
    \item We enable DP deep learning with more than 1B trainable parameters for the first time. E.g. we are the first to train the full GPT2-XL, ViT-Gigantic, ViT-10B and GPT-100B with DP.
    \item We will open-source our codebase that automatically applies DP-ZeRO for general tasks (e.g. classification and language understanding), general network architectures (e.g. ResNet, ViT, GPT), and general distributed solutions (including DeepSpeed and FSDP), with one line of code change.
\end{enumerate}

\begin{figure}[!htb]
\vspace{-0.5cm}
    \centering
    \includegraphics[width=0.4\linewidth]{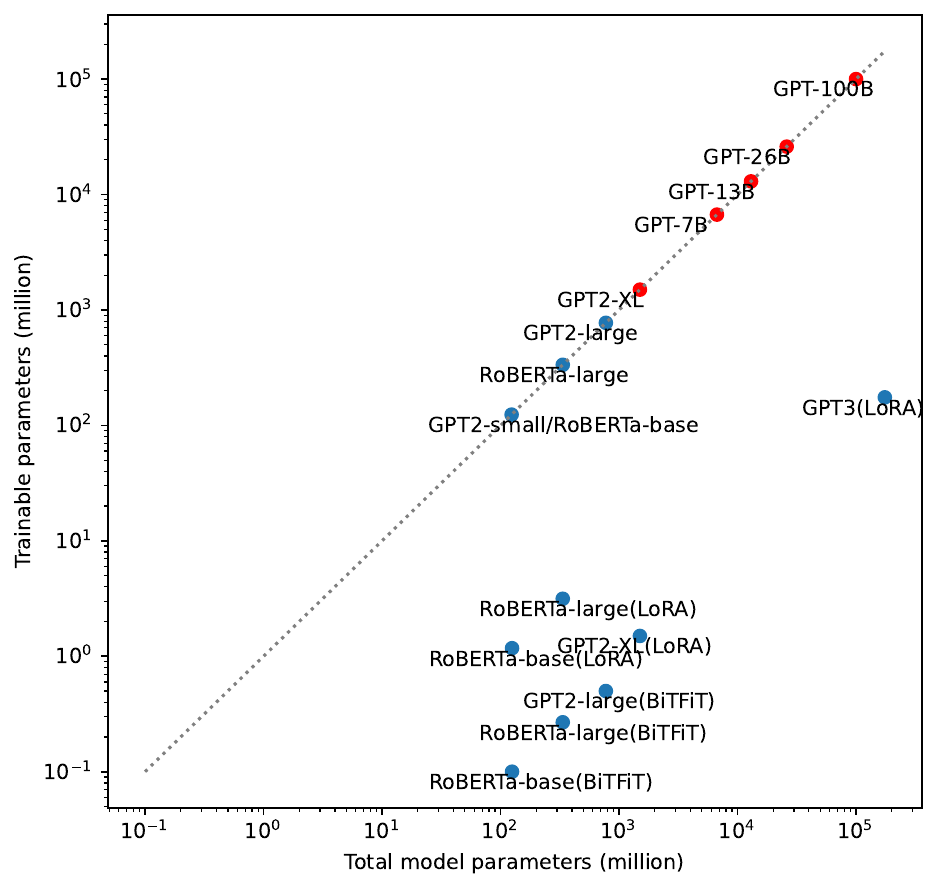}
    \includegraphics[width=0.4\linewidth]{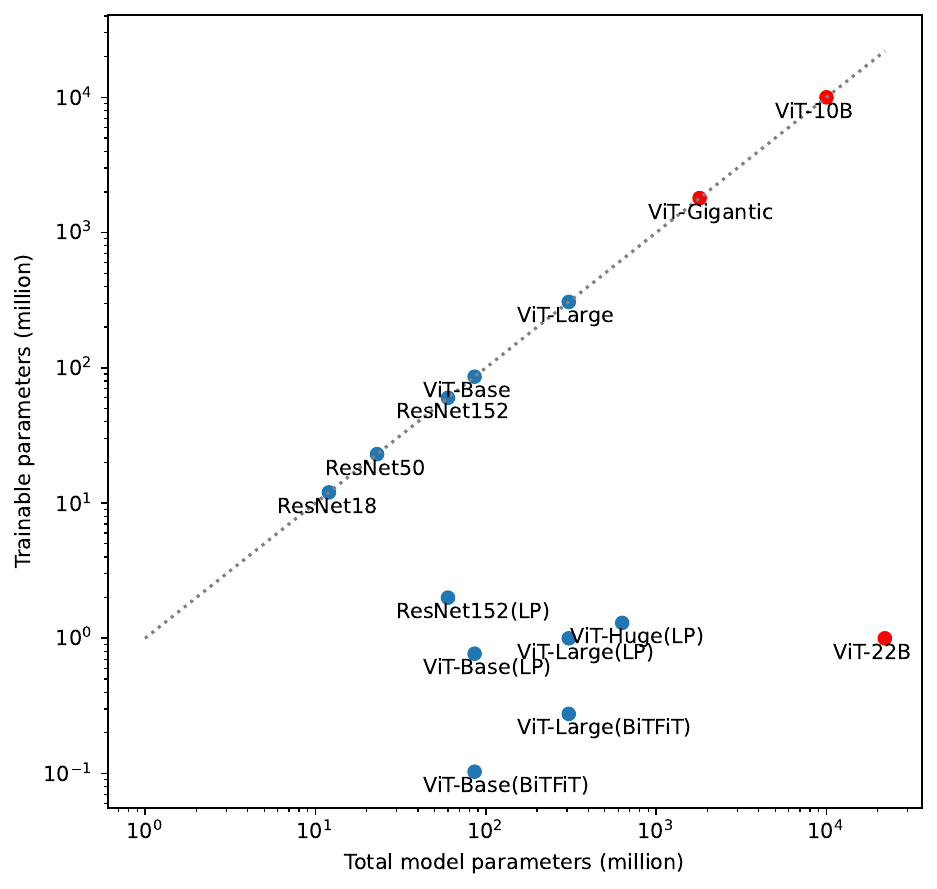}
    \vspace{-0.3cm}
    \caption{Total/trainable parameters of existing DP models \textcolor{blue}{(blue)} and ours by DP-ZeRO \textcolor{red}{(red)}.}
    \label{fig:my_label}
    \vspace{-0.5cm}
\end{figure}

% \begin{itemize}
%     \item DP: clipping algorithm, clipping style, mixed-precision
%     \item efficiency: parameter efficiency, computation efficiency (time and memory),
%     \item distributed learning: communication efficiency, partition stages,
% \end{itemize}

\section{Preliminary}
\subsection{Differential privacy}
\label{sec:DP prelim}
DP provides a formal privacy guarantee, making it difficult to extract any information from training data. The privacy guarantee is characterized by $(\epsilon,\delta)$-DP in \Cref{def:DP}, with smaller $(\epsilon,\delta)$ representing lower privacy risk.

\begin{definition}[\cite{dwork2006calibrating}]\label{def:DP}
A randomized algorithm $M$ is $ (\varepsilon, \delta)$-DP if, for any two neighboring datasets $S,S^{\prime}$ that differ by one sample and for any event $E$, we have $\mathbb{P}[M(S) \in E] \leqslant \mathrm{e}^{\varepsilon} \mathbb{P}\left[M\left(S^{\prime}\right) \in E\right]+\delta.$
\end{definition}

In DP deep learning, the gradients are made private by post-processing through per-sample gradient clipping and random noising:
\begin{align}
\text{private gradient: }\G_{[m]}:=\sum_i C_i(R_m) \g_{[m],i}+\sigma_\text{DP}\|[R_1,\cdots,R_M]\|\cdot\mathcal{N}(0,\I),
\label{eq:private grad}
\end{align}
Here the gradient of all trainable parameters is partitioned into $M$ groups, i.e. $\g_{[m],i}$ is the $i$-th per-sample gradient of the $m$-th group's parameters, where $m\in \{1\cdots M\}$ is the group index. $C_i$ is the per-sample gradient clipping factor so that $\|C_i \g_{[m],i}\|\leq R_m$ and $R_m$ is the clipping threshold. That is, DP optimization is enabled when the standard optimizers such as stochastic gradient descent (SGD) and Adam \citep{kingma2014adam} update the trainable parameters with the private gradient, instead of the standard gradient $\sum_i \g_i$.

\paragraph{Mathematical gradient partition}
In \eqref{eq:private grad}, the trainable parameters and their gradients are mathematically partitioned into $M$ groups, e.g. in all-layer clipping, all parameters form one group ($M=1$) \cite{abadi2016deep}; in layer-wise clipping \cite{mcmahan2018learning,anonymous2023exploring}, each layer's parameters form a group ($M=$ number of layers). 
% Other examples include per-GPU \cite{anonymous2023exploring}, parameter-wise \cite{opacus}, and uniform grouping.
Empirical evidence and theoretical analysis show that different partitions have the same training speed, though a finer partition (e.g. layer-wise) has lighter memory footprint\footnote{We note that DP optimization under different group-wise clippings can have the same computation and communication efficiency (under the BK algorithm), with or without ZeRO.
% However, current implementations of distributed learning, e.g. ZeRO, may encounter different levels of difficulty when adapting to different group-wise clippings. We extend this discussion in appendix
}.

\paragraph{Per-sample gradient clipping}
In \eqref{eq:private grad}, a number of clipping functions $C_i=C(\|\g_i\|;R)$ are available. Most works \cite{abadi2016deep, li2021large,yu2021differentially} use the vanilla clipping $C_i=\min(R/\|\g_i\|,1)$. Recently,  \cite{bu2022automatic,yang2022normalized} advocate the automatic clipping $C_i=1/(\|\g_i\|+0.01)$ which is hyperparameter-free and comparably accurate.
Note if $C_i\equiv 1$, then the clipped gradient reduces to the standard gradient. % but the clipped gradient is not guaranteed to have bounded magnitude.
The main overhead of DP optimization is the computation of per-sample gradient norms. On a single GPU, the mixed ghost clipping \cite{bu2022differentially} has reduced the time complexity to $<10\%$ on large models like GPT2.

\paragraph{Privacy accounting}
In \eqref{eq:private grad}, adding Gaussian noise to the clipped gradient protects the privacy that is quantifiable by the privacy accounting theory \cite{abadi2016deep,bu2020deep,dong2019gaussian,zhu2021optimal,gopi2021numerical,koskela2020computing}. The privacy guarantee is increasing in the noise level $\sigma_\text{DP}$, independent of $R_m$, learning rate, clipping function and model architectures, with $\sigma_\text{DP}=0$ leading to $\epsilon=\infty$ (non-private). 

\subsection{Zero Redundancy Optimizer (ZeRO)}
\subsubsection{Parallel computing}
Parallel computing is necessary to train large-scale models and is critical to the optimization efficiency. For models that fit in a single GPU, \underline{data parallelism (DataP)} can be used to speed up the training by partition the mini-batch of samples into multiple micro-batches. Then, each GPU (holding a full copy of parameters) executes the forward and backward propagation of one micro-batch, from which the parameter gradients are generated and averaged across GPUs to update the trainable parameters. However, for models that do not fit in a single GPU, the model parameters need to be sharded by alternative solutions such as ZeRO \cite{rajbhandari2020zero}, \underline{model parallelism (ModelP)} and \underline{pipeline parallelism (PipeP)}.

ModelP partitions a model vertically, e.g. using 3 GPUs to store the parameters of one layer. As a consequence, ModelP does not scale efficiently beyond a single node due to fine-grained computation and expensive communication between layers. Implementation-wise, ModelP frameworks usually require heavy code integration that may not be generalizable in model architectures. In contrast, PipeP partitions a model horizontally across layers, e.g. storing 3 layers in each GPU. Each GPU deals with all micro-batches sequentially, though PipeP can be inefficient due to the pipeline bubble, which is overcome by ZeRO \cite{rajbhandari2020zero}. ZeRO is an advanced data parallel method that eliminates memory redundancies during the training, and improves the training speed and communication volume proportionally to the number of GPUs. Unlike basic DataP, ZeRO partitions a model's states across GPUs and gather/reduce in a just-in-time manner, thus sustaining the high efficiency of very large model training. Notice that ZeRO can work compatibly with ModelP and optionally offload the model states to CPUs \cite{ren2021zero,rajbhandari2021zero}.

\subsubsection{Model state partition}
\label{sec:model partition}
A major part of the training memory is consumed by the model states\footnote{Another important part of memory consumption is the batch-size-related variables such as the activation tensors, which is instantiated during the forward propagation and independent to DP (which modifies the gradient during the back-propagation).}. ZeRO has three stages (ZeRO1/2/3) that partition these model states by different levels, with lower level being faster but more memory costly. For instance, in Table 2 of \citep{rajbhandari2020zero}, ZeRO1/2/3 at most train 7.6/14.4/128B models on 64 V100 GPUs.

We take an example of mixed-precision Adam optimizer to train a model  with $\Psi_\text{model}$ parameters, which maintains a master copy (fp32) of optimizer states 
% \footnote{The memory cost of fp32 gradients is ignored, because they are instantiated and then deleted on one layer at a time, hence only existing temporarily.}
, and the half-precision parameters and gradients.

\paragraph{Optimizer state partition
% ($P_\text{optim}$)
}
The optimizer states are the (master) parameters, variance and momentum, each taking $4\Psi_\text{model}$ memory. ZeRO1 only applies the optimizer state partition and updates the parameters locally, reducing the memory cost of model states from $16\Psi_\text{model}$ for basic DataP to $(4+\frac{12}{N_d})\Psi_\text{model}$ at each of $N_d$ GPUs.

\begin{wrapfigure}{r}{0.53\linewidth}
    \vspace{-1.3cm}
    \centering
    \includegraphics[width=\linewidth]{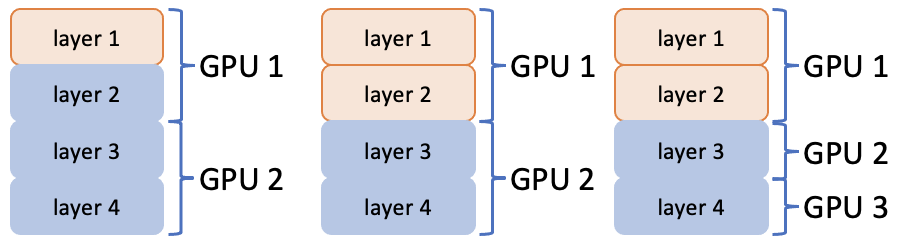}
    \caption{Mathematical (left two, same GPU allocation, different accuracy) and hardware (right two, different GPU allocation, same accuracy) gradient partition. Orange and blue are gradient groups $\g_{[1]}$ and $\g_{[2]}$ in DP optimization \eqref{eq:private grad}.}
    \label{fig:grad partition}
    \vspace{-1cm}
\end{wrapfigure}

\paragraph{Hardware gradient partition 
% ($P_\text{grad}$)
}
In addition to ZeRO1, during the back-propagation, ZeRO2 (and ZeRO3) further partitions the $2\Psi_\text{model}$ gradients into different GPUs, reducing the memory cost to $(2+\frac{14}{N_d})\Psi_\text{model}$ at each GPU. \Cref{fig:grad partition} illustrates the difference between the hardware partition and the mathematical partition in \Cref{sec:DP prelim}.

\paragraph{Parameter partition
% ($P_\text{param}$)
}
In addition to ZeRO2, ZeRO3 also partitions the $2\Psi_\text{model}$ fp16 parameters,  further reducing the memory cost to $\frac{16}{N_d}\Psi_\text{model}$ at each GPU.

\subsection{Mixed-precision training}
\label{sec:mix precision}
Mixed-precision training \citep{micikevicius2018mixed} performs the forward and backward propagation on the half precision (fp16 or bf16) parameters, activations, and gradients, while performing the model update in full precision (fp32). Compared to the full-precision training, it is capable of saving the memory by $\approx 50\%$ and accelerating the computation by $\approx 20\%$. We note that fp16 has better precision but a limited range: its representable numbers are among $10^{-8}\sim 10^5$, and vice versa for bf16. Therefore, loss scaling is necessary when using fp16 to prevent small gradients from being rounded to zero, thereby preserving the model's accuracy (see \Cref{app:loss scaling} for details). In contrast, bf16 usually does not need loss scaling since it has the same range as fp32. However, bf16 is not as widely supported as fp16, e.g. only available on NVIDIA Ampere GPUs or above \citep{NVIDIA}.

\iffalse
\begin{table}[!htb]
\centering
\begin{tabular}{c|c|c|c}
& model & \# param($\Psi_\text{model}$)& \# trainable param($\Psi_\text{train}$) \\
\cite{anonymous2023exploring}&GPT3 &175B&151M \\
\cite{bu2022differentially}&GPT2-large &774M&774M \\
BiTFiT&GPT2-large &774M&0.5M \\\hline
this work&ViT-Gigantic &1.84B&1.84B \\
this work&GPT2-XL &1.56B&1.56B \\
this work&GPT3 &175B&175B \\
\end{tabular}
\caption{Caption}
\label{tab:my_label}
\end{table}
\fi

\iffalse
\subsection{Fused operations and time-memory tradeoff}
% https://pytorch.org/tutorials/recipes/recipes/tuning_guide.html#fuse-pointwise-operations
During the back-propagation, pointwise operations like tensor multiplication and math functions can be time-consuming due to the memory access time and the kernel launch time. This is because each operation launches a separate kernel to read data from the memory, compute and save the results back to the memory. Alternatively, multiple operations can be fused into a single kernel that access the memory only once. However, a significant time-memory tradeoff may be observed: while fused operations are generally faster than their non-fused counter-parts, the peak memory cost may be much higher for deep neural networks, say 30\% for ViT-Gigantic (2B). Given that the saved memory allows one to use larger batch size, which translates to faster training, it is likely that the non-fused back-propagation of very large models can out-speed the fused version.
\fi

\section{Differentially private ZeRO}

\subsection{Algorithm}
Our DP-ZeRO algorithm introduces the per-sample gradient clipping and noising to the standard ZeRO \citep{rajbhandari2020zero}, while maintaining the efficiency. At high level, an iteration of ZeRO consists of the following steps:
$$\resizebox{.95\hsize}{!}{$\left(\colorbox{Plum!20}{    all-gather}\to\colorbox{LimeGreen!20}{forward}\to\colorbox{Plum!20}{all-gather}\to\colorbox{LimeGreen!20}{backward}\to\colorbox{Plum!20}{reduce}\right)^{\times L \text{ layers}}\to\colorbox{LimeGreen!20}{update(SGD/Adam/...)}$}$$
The operations in \colorbox{Plum!20}{purple} are global and require communication among GPUs, whereas the operations in \colorbox{LimeGreen!20}{green} are locally computed within each GPU. In particular, DP optimization is only different from the standard optimization in the back-propagation, which can be decomposed into
$$\colorbox{LimeGreen!20}{backward}= (\text{output gradient}\to\text{clipping factor}\to\text{parameter gradient}\to\text{noising})$$

% \begin{figure}[!htb]
%     \centering
%     \includegraphics[width=0.6\linewidth]{local_global.png}
%     \vspace{-0.2cm}
%     \caption{Back-propagation of DP-ZeRO. The red dashed line is the additional computation from DP. %Note that ZeRO1/2 do not need to all-gather the parameters before back-propagation. Also, in (non-DP) ZeRO, the clipping factor is always treated as 1.
%     }
%     \label{fig:local global}
%     \vspace{-0.5cm}
% \end{figure}

\begin{figure}[!htb]
    \centering
\vspace{-0.4cm}
    \includegraphics[width=0.6\linewidth]{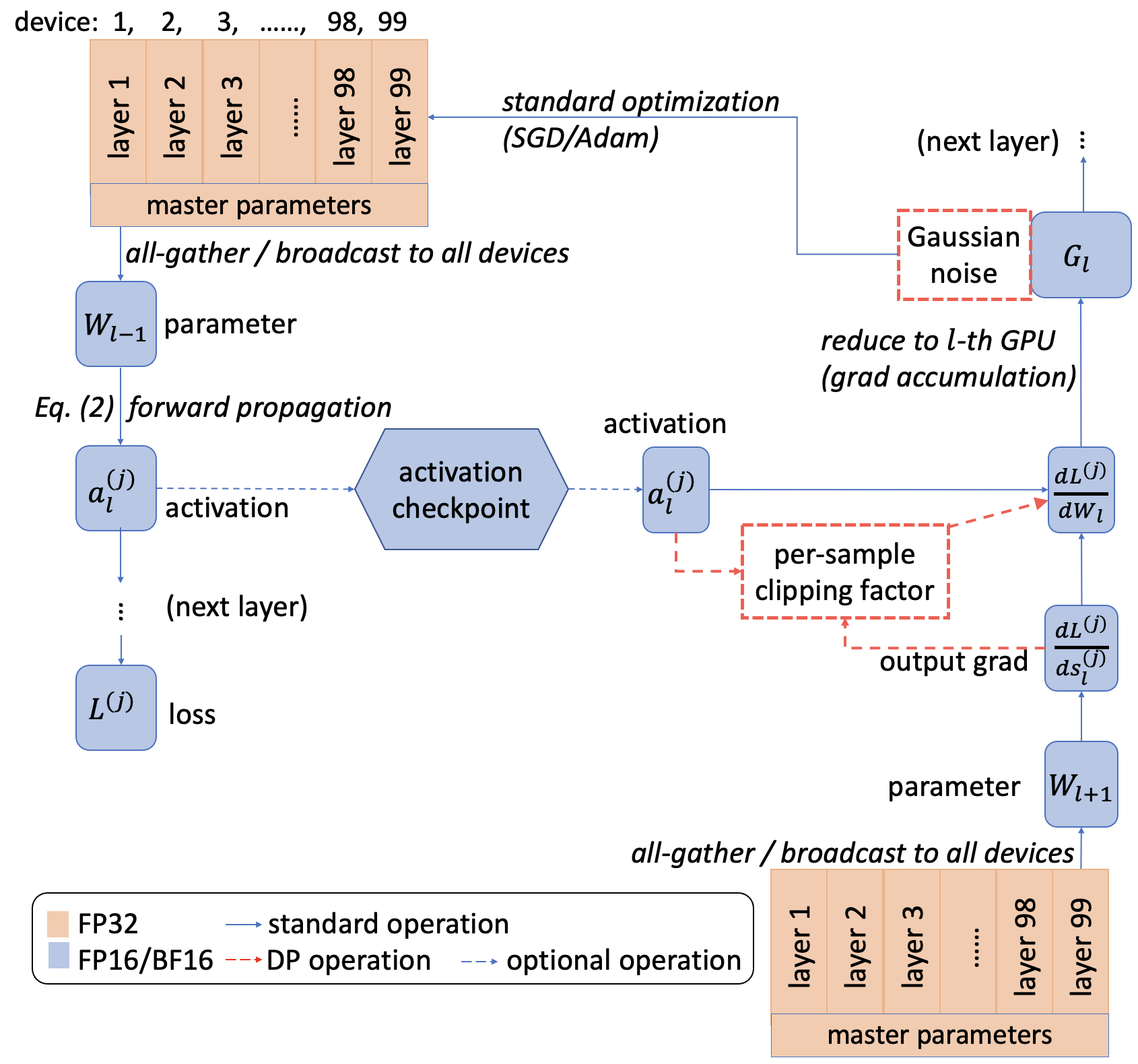}
    \vspace{-0.2cm}
    \caption{Algorithm of DP-ZeRO with mixed-precision training.}
    \label{fig:flowchart}
    % \vspace{-0.5cm}
\end{figure}

To give more details, we consider training a neural network of linear layers using $N_d$ GPUs. We emphasize that the following procedure is sufficiently generic to cover other layer types, such as convolution, embedding, normalization, and so on, which are all supported by DP-ZeRO. The full algorithm is depicted in \Cref{fig:flowchart}, where we denote the $j$-th micro-batched variables like $\a_l^{(j)}$, for $1\leq j\leq N_d$.

The forward propagation of DP optimization is the same as that of the standard optimization: 
\begin{align}
\s_{l}=\a_{l}\W_{l}+\b_{l}, \a_{l+1}=\phi_l(\s_{l})
.
\label{eq:forward}
\end{align}
At the $l$-th layer, $\a_l\in\R^{BT_l d_l}$ is the layer's input, also known as the activation, $\s_l\in\R^{BT_l p_l}$ is the layer's output, $\W_l\in\R^{d_l p_l}$ is the weight, $\b_l\in\R^{p_l}$ is the bias, and $\phi_l$ is any inter-layer operation such as ReLU, tanh or pooling. We denote $B$ as the physical micro-batch size\footnote{The micro-batch size $B$ is the number of samples processed by each GPU, which determines the time and memory efficiency, but not the performance. The logical batch size that determines the performance is $B\times N_d\times \text{gradient\_accumulation\_steps}$. 
% For example, if the batch size is 1024 and we use 8 GPUs, we can feed 64 samples (physical batch size) for one step, and only updates the model when we accumulates the gradient for two steps.
} and $T_l$ as the hidden feature dimension (e.g. sentence length or number of pixels). %In what follows, we focus on the layer-wise clipping for the simplicity of presentation, though other clipping styles (including the all-layer / flat clipping) are supported by DP-ZeRO.
During the forward propagation, the activations $\{\a_l\}$ are computed and stored in the computation graph, and the loss $L=\sum_i L_i$ is derived, where $L_i$ is the per-sample losses. During the back-propagation, the output gradient is first computed based on the previous layer.
$$\frac{\partial L}{\partial \s_{l}}=\frac{\partial L}{\partial \s_{l+1}}\frac{\partial \s_{l+1}}{\partial \a_{l+1}}\circ\frac{\partial \a_{l+1}}{\partial \s_{l}}=\frac{\partial L}{\partial \s_{l+1}}\W_{l+1}\circ\phi_l'(\s_{l}),$$
in which $\circ$ is element-wise multiplication. Specifically, the use of parameter $\W_{l+1}$ necessitates the all-gather operation when the model is partitioned into multiple GPUs, which is not needed in single GPU training.
Next, the activation $\a_l$ is used together with $\frac{\partial L}{\partial \s_{l}}$ to compute the parameter gradient:
\begin{align*}
% \text{Standard (non-DP): }&\quad\quad \frac{\partial L}{\partial \W_{l}}=\frac{\partial \sum_i L_i}{\partial \W_{l}}=\a_l^\top\frac{\partial L}{\partial \s_{l}},&\\
\text{DP gradient: }&\frac{\partial \sum_i C_i L_i}{\partial \W_{l}}+\sigma \mathcal{N}(0,\mathbf{I})=\a_l^\top\text{diag}(C_1,\cdots,C_B)\frac{\partial L}{\partial \s_{l}}+\sigma \mathcal{N}(0,\mathbf{I}).&
\end{align*}
Note the standard gradient can be viewed as $C_i=1, \sigma=0$. Here the per-sample gradient norm (or the clipping factor $C_i$) can be computed at small cost, as we have discussed in \Cref{sec:intro}. 
% Finally, the Gaussian noise in \eqref{eq:private grad} is can be easily injected before any optimizer updates the parameters.

\subsection{Time efficiency of DP-ZeRO}
\label{sec:time efficiency}
The time efficiency of DP-ZeRO consists of two parts: the local computation (including forward and backward propagation) and the global communication (including intra-node and inter-node communication). Given that the only difference between DP-ZeRO and ZeRO is the back-propagation, we claim that DP-ZeRO could enjoy high efficiency on-par with the standard ZeRO when (I) DP back-propagation exhibits a time efficiency comparable to the standard, similar to the single GPU training, and/or (II) the time efficiency of the parts other than back-propagation is not
insignificant. We give the time of each part of DP-ZeRO in \eqref{eq:relative DP speed} to illustrate our claim. 

\vspace{-0.5cm}
\begin{align}
\frac{\text{DP-ZeRO speed}}{\text{standard ZeRO speed}}
=\frac{\textcolor{blue}{\text{back-propagation}}+\textcolor{Green}{\text{forward propagation}}+\textcolor{Orange}{\text{communication}}}{\textcolor{Red}{\text{DP back-propagation}}+\textcolor{Green}{\text{forward propagation}}+\textcolor{Orange}{\text{communication}}}
\label{eq:relative DP speed}
\end{align}
To be explicit, we summarize the time complexity in \Cref{tab:complexity full finetune} and refer to \Cref{app:time complexity} for details.

\begin{table}[!htb]
	\centering
\caption{Time complexity of one iteration in distributed learning\protect\footnotemark. We denote $\Psi_\text{train}$ to be the number of trainable parameters ($\Psi_\text{train}=\Psi_\text{model}$ in full parameter training), and define $B,T$ below \eqref{eq:forward}.}
	\vspace{-0.2cm}
	\resizebox{0.99\linewidth}{!}{
	\begin{tabular}{|c|c|c|c|c|c|c|}
	\hline
	& \multicolumn{2}{|c|}{forward propagation}&\multicolumn{3}{|c|}{back-propagation}&\multirow{2}{*}{communication}
		\\\cline{2-6}
		&activation&attention&output grad &param grad &DP clip and noise&\\\hline
		complexity & $2BT\Psi_\text{model}$&$O(BT^2)$& $2BT\Psi_\text{model}$& $2BT\Psi_\text{train}$&$ 0.666BT\Psi_\text{train}$&$O(\Psi_\text{model})$\\\hline
	\end{tabular}
	}
	\label{tab:complexity full finetune}
\end{table}
\footnotetext{Here 0.666 is figurative and dependent on settings. Notice that $\Psi_\text{train}\ll\Psi_\text{model}$ when most parameters are frozen.}

% We can see that the choice of hyperparameters (e.g. learning rate), DP clipping style (e.g. layer-wise or all-layer) and training precision do not influence the relative training speed of DP-ZeRO. 

In what follows, we analyze the absolute and relative speed (to standard ZeRO) of DP-ZeRO under important settings.

\iffalse
\begin{table}[!htb]
	\centering
	\begin{tabular}{c|c|c}
		&forward&backward\\\hline
		DDP & --- (0) & all-reduce ($2\Psi_\text{train}$)
		\\
		ZeRO 1 & --- (0)&reduce-scatter + all-gather + (update) ($\Psi_\text{train}+\Psi_\text{train}$)
		\\
		ZeRO 2 & --- (0)&reduce-scatter + (update) + all-gather ($\Psi_\text{train}+\Psi_\text{train}$)
		\\
		ZeRO 3 & all-gather ($\Psi_\text{model}$)&
		all-gather + reduce-scatter + (update) ($\Psi_\text{model}+\Psi_\text{train}$)   
	\end{tabular}
	\caption{scatter if $P_{os}$}
	\label{tab:my_label}
\end{table}
\fi

\subsubsection{Number of computation devices}
\label{sec:number GPU}
% The communication cost is determined by the communication efficiency and communication volume.
% \paragraph{Communication efficiency }
When scaling from one GPU (zero communication) to one node (multiple GPUs) and to multiple nodes, the communication efficiency decreases sub-linearly (\textcolor{Orange}{$\uparrow$communication}). On a single node, multiple GPUs can communicate using the high-speed intra-node connections such as NVLink/NVSwitch \citep{foley2017ultra,ishii2018nvswitch}. On multiple nodes, which are necessary for large models, the inter-node connections are $3\sim 24\times$ slower than the intra-node connections \citep{li2019evaluating,zhang2022mics}. In short, DP-ZeRO can be as fast as ZeRO by \eqref{eq:relative DP speed} when multiple nodes are employed.

\subsubsection{Memory-efficient distributed learning}
The communication volume is specific to different distributed algorithms, most of which trade the communication or speed for memory, in order to feasibly train very large models. For example, ZeRO3 (but not ZeRO1/2) needs to all-gather the sharded parameters at each iteration, hence suffering from 50\% extra communication volume (\textcolor{Orange}{$\uparrow$communication}). Another example is the activation check-pointing (also known as gradient check-pointint \cite{chen2016training}), where a second forward propagation re-computes the expensive activations during back-propagation, though at a 33\% slower speed (\textcolor{Green}{$\uparrow$forward propagation}). These techniques improve the relative speed of DP-ZeRO but worsens the absolute speed.
% \footnote{In standard distribute learning, the gradient compression methods such as signSGD \citep{bernstein2018signsgd}, quantization \citep{alistarh2017qsgd}, and sparsification \citep{aji2017sparse} can significantly the communication volume, at the risk of performance degradation. However, extra efforts are needed to adapt these methods to DP optimization.}

\subsubsection{Parameter efficient fine-tuning}
\label{sec:PEFT}
Parameter efficient fine-tuning (PEFT), such as LoRA \citep{hu2021lora}, Adapter \citep{houlsby2019parameter}, and BiTFiT \citep{zaken2022bitfit}, optimizes a small fraction (e.g. $\Psi_\text{train}=0.1\%\Psi_\text{model}$) of model parameters and thus boosts the efficiency of back-propagation and communication (\textcolor{Red}{$\downarrow$DP back-propagation} \textcolor{blue}{$\downarrow$back-propagation} \textcolor{Orange}{$\downarrow$communication}). Consequently, (I) the communication volume of the gradient can be reduced possibly by $1000\times$; (II) the local computation can accelerate by 50\% \citep{hu2021lora,bu2022dpbitfit}, which can be seen by treating $\Psi_\text{train}$ in \Cref{tab:complexity full finetune} as almost zero; (III) the memory cost is saved on the non-trainable layers% by not storing the gradients and optimizer states
, which translates to larger batch size and faster computation. Hence, both relative and absolute speed of DP-ZeRO improve using PEFT.

%In \cite{rajbhandari2020zero}, only full fine-tuning is considered, and the communication cost is DDP=ZeRO1=ZeRO2=2/3 ZeRO3. However, PEFT changes this to $\text{ZeRO1}=\text{ZeRO2}\ll\text{ZeRO3}$.

%Equivalently, the communication cost if $\Psi_\text{model}=100\Psi_\text{train}$,
%$$\text{ZeRO3(full)}=1.49\text{ZeRO3(PEFT)}=1.5\text{ZeRO2(full)}=1.5\text{ZeRO1(full)}=150\text{ZeRO2(PEFT)}=150\text{ZeRO1(PEFT)}$$

\iffalse
\begin{align*}
	\frac{\text{Time}_\text{comp}\cdot[\frac{2}{3}+\frac{1}{3}\mathbb{I}_\text{\{PEFT=False\}}(1+0.5\mathbb{I}_\text{\{DP=True\}})]+\text{Time}_\text{comm}\cdot[\mathbb{I}_{\{\text{ZeRO}= 3\}}(1-0.5\mathbb{I}_{\{\text{PEFT=True}\}})+\mathbb{I}_{\{\text{PEFT=True}\}}]}{\mathbb{I}_{\{\text{ZeRO}\neq 3 \text{ or FiTOneGPU=True}\}}}
\end{align*}

ZeRO=0,1,2,3; haven't consider mixed-precision/one-bit adam/activation checkpointing; FiTOneGPU means parameter
\fi

\subsection{Memory efficiency of DP-ZeRO}
We claim that DP-ZeRO is as memory efficient as the standard ZeRO, similar to the single GPU training, when we use (I) the mixed ghost norm trick \cite{bu2022scalable,bu2022differentially}, instead of GhostClip \citep{goodfellow2015efficient,li2021large} or per-sample gradient instantiation \citep{opacus}; (II) the layer-wise clipping style instead of the all-layer clipping, so that the book-keeping \citep{bu2022differentially} does not store all output gradients; (III) a large number of GPUs so that the micro-batch size $B$ (i.e. per-GPU batch size) is small: specially, when 
% the logical batch size equals $N_d\times \text{gradient\_accumulation\_steps}$, we have 
$B=1$ in the gradient accumulation, the per-sample gradient is free. We empirically verify our claim in \Cref{fig:generality,fig:multiple stages}.

\subsection{Mixed-precision training with DP}

\begin{wrapfigure}{r}{0.33\linewidth}
    \vspace{-0.9cm}
    \includegraphics[width=\linewidth]{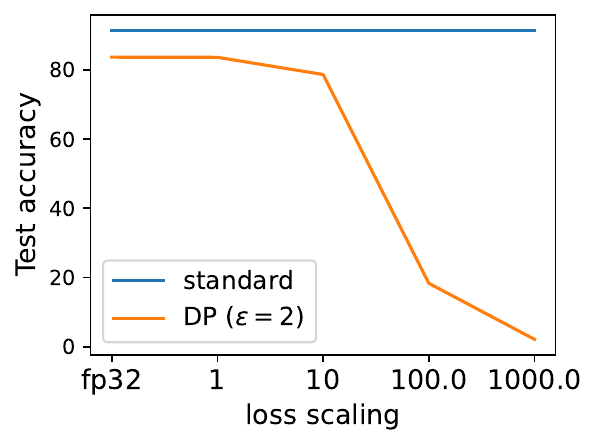}
    \vspace{-0.8cm}
    \caption{Accuracy of mixed-precision training with loss scaling. ViT-large on CIFAR100.}
    \label{fig:DP loss scaling}
    \vspace{-0.2cm}
\end{wrapfigure}

We now analyze the intricacy in mixed-precision training with DP, which is not unique to DP-ZeRO but present in the general DP optimization. We emphasize that the per-sample gradient clipping already plays the role of scaling, and hence DP mixed-precision training must not use loss scaling, as illustrated in \Cref{tab:loss scale}. Specifically, in standard mixed-precision training, there are two steps of scaling: (I) scaling up the loss ${L}_i$ by $10^3\sim 10^9$ (and consequently the output gradient $\frac{\partial {L}_i}{\partial \s}$ as well as $\frac{\partial {L}_i}{\partial \W}$) before the back-propagation, to prevent the underflow where fp16 gradient is too small to be distinguished from 0, and (II) scaling down the parameter gradient $\frac{\partial {L}_i}{\partial \W}$, by the same factor, after the back-propagation to recover the correct magnitude of gradient. However, in DP mixed-precision training, scaling up the loss may cause overflow, while scaling down the gradient incorrectly over-shrinks the gradient and worsens the performance. See \Cref{fig:DP loss scaling} for a real example. We explain this intricacy step-by-step in \Cref{app:loss scaling}.

\begin{table}[!htb]
\centering
\caption{Illustration of \textcolor{red}{overflow} and \textcolor{Green}{underflow} issues during mixed-precision training (%fp16, with 
ghost norm).}
\resizebox{\linewidth}{!}{
\begin{tabular}{|c|c|c|cc|c|c|c|}
\hline
\multirow{2}{*}{loss scale=$10^3$} & \multirow{2}{*}{activation $\a_l$} & \multirow{2}{*}{output grad $\frac{\partial L}{\partial \s_l}$} & \multicolumn{2}{c|}{per-sample grad norm}                           & \multirow{2}{*}{clipping factor} & param grad & param grad \\ \cline{4-5}
   &                             &          (scaled)            & \multicolumn{1}{c|}{$\text{vec}(\a_l\a_l^\top)$}              & $\text{vec}(\frac{\partial L}{\partial \s_l}\frac{\partial L}{\partial \s_l}^\top)$                && (not scaled down)      &(if scaled down)\\ \hline
standard w/o scaling               & $10^{-3}\sim10^2$              & $\textcolor{Green}{10^{-8}}\sim10^{1}$            & \multicolumn{1}{c|}{N/A}             & N/A               & 1    
& $10^{-7}\sim10^{1}$                        &$10^{-7}\sim10^{1}$            \\ \hline
standard w/ scaling                & $10^{-3}\sim10^2$              & $10^{-5}\sim10^{4}$            & \multicolumn{1}{c|}{N/A}             & N/A               & 1                          & $10^{-4}\sim10^{4}$                        & $10^{-7}\sim10^{1}$                             \\ \hline
DP w/o scaling                     & $10^{-3}\sim10^2$              & $\textcolor{Green}{10^{-8}}\sim10^{1}$            & \multicolumn{1}{c|}{$10^{2}\sim10^{3}$} & $10^{-6}\sim10^{0}$ & $10^{-3}\sim10^{2}$             & $10^{-7}\sim10^{1}$                        &$10^{-7}\sim10^{1}$                  \\ \hline
DP w/ scaling                      & $10^{-3}\sim10^2$              & $10^{-5}\sim10^{4}$            & \multicolumn{1}{c|}{$10^{2}\sim10^{3}$} & $10^{0}\sim\textcolor{red}{10^{6}}$   & $10^{-6}\sim10^{-1}$            & $10^{-7}\sim10^{1}$                        &$\textcolor{red}{10^{-10}\sim10^{-2}}$                   \\ \hline
\end{tabular}
}
\vspace{-0.2cm}
\label{tab:loss scale}
\end{table}

\section{Empirical performance of DP-ZeRO}
We evaluate DP-ZeRO on five aspects: model architectures, efficiency, scalability, compatibility with various distributed learning and DP techniques. We use DP-ZeRO to train ResNet \citep{he2016deep}, ViT \citep{dosovitskiy2020image,zhai2022scaling} and GPT \citep{radford2019language,brown2020language}, which are workhorses in computer vision and language tasks. We measure the time and memory efficiency of DP-ZeRO under settings such as PEFT and multiple precision formats (fp32 or fp16/bf16). We evaluate the scalability of DP-ZeRO in terms of number of GPUs and number of model parameters. Our experiments scale from single node (8 GPUs) to multiple nodes, up to 256 GPUs, and train models up to 100B trainable parameters. Moreover, DP-ZeRO is compatible with mainstream implementations of ZeRO\footnote{DP-ZeRO is implemented on DeepSpeed (supporting ZeRO1/2/3), FSDP \citep{FSDP} (supporting ZeRO3), MiCS \citep{zhang2022mics} (supporting ZeRO2/3), and any distributed optimizers supported on them.} and with different clipping styles, clipping functions, privacy accountants, and so on.
We leave the experimental details in \Cref{app:settings}. By default, we use AdamW, mixed-precision training, layer-wise clipping style, $B=4$, and A100 GPU with 40GB memory, unless otherwise stated.

\subsection{Generality of DP-ZeRO}
DP-ZeRO is generally applicable to different neural network architectures, clipping styles, and precision formats. We test DP-ZeRO1 on single node and observe that different clipping styles are equally fast, but layer-wise clipping is more memory efficient than all-layer clipping. 
% and observe a pattern that is consistent with the single-GPU DP optimization
Comparing to the standard ZeRO, our DP-ZeRO enjoys almost the same speed and memory efficiency, while the gap will be further closed as we move to more advanced stages of ZeRO.
\begin{figure}[!htb]
    \centering
    \includegraphics[width=0.16\linewidth]{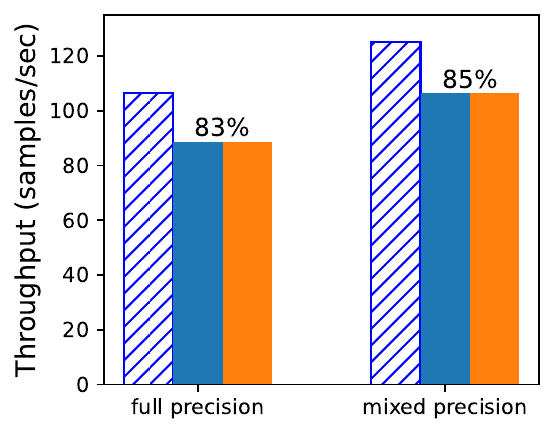}
\includegraphics[width=0.16\linewidth]{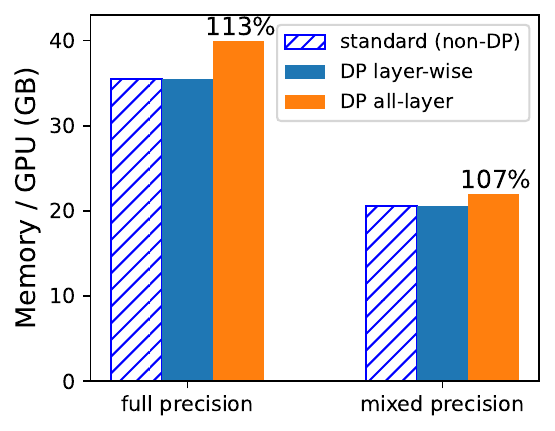}
    \includegraphics[width=0.16\linewidth]{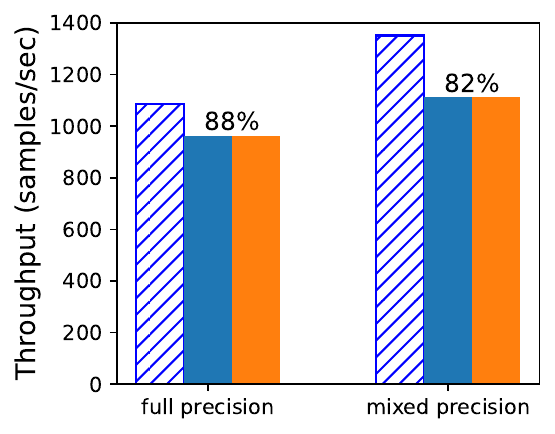}
\includegraphics[width=0.16\linewidth]{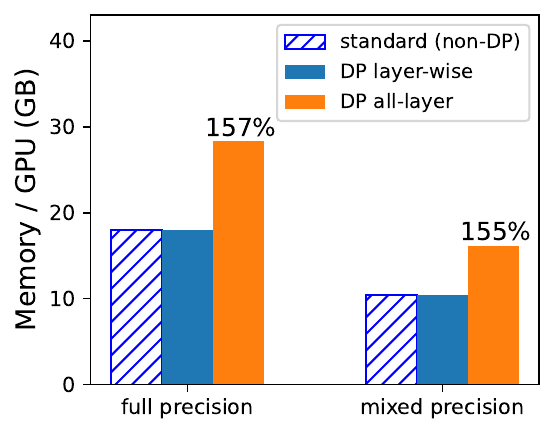}    
\includegraphics[width=0.16\linewidth]{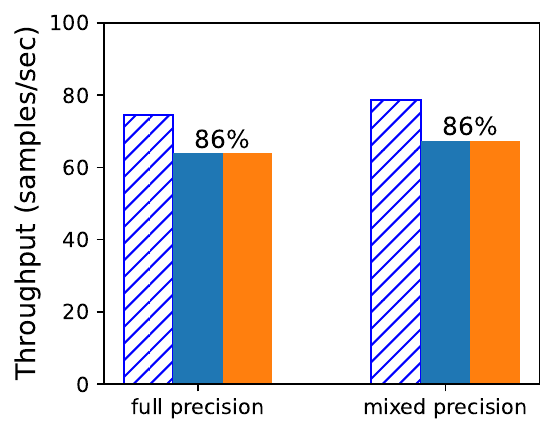}
\includegraphics[width=0.16\linewidth]{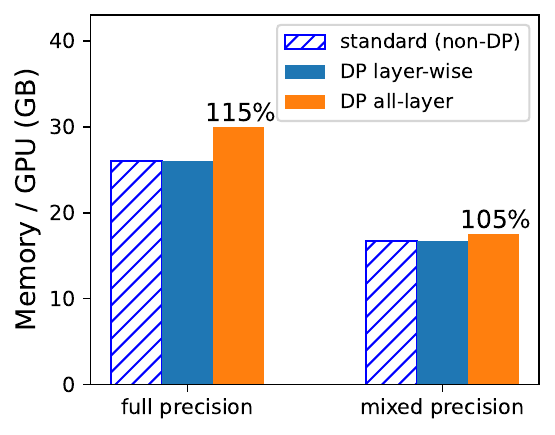} 
\vspace{-0.2cm}
\caption{Efficiency on ViT-Gigantic (left, 1.8B), ResNet152 (middle) and GPT2-XL (right, 1.5B).}
    \label{fig:generality}
\end{figure}

\vspace{-0.8cm}
\subsection{Lighter training of DP-ZeRO}
\begin{wrapfigure}{r}{0.35\linewidth}
    \vspace{-0.8cm}
\centering
    \includegraphics[width=0.38\linewidth]{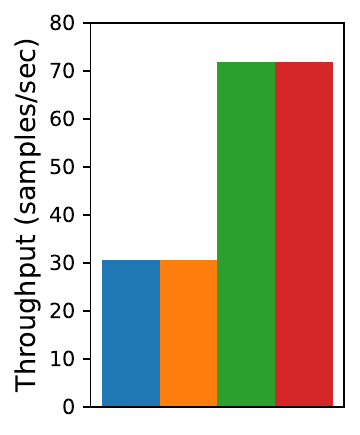}
    \includegraphics[width=0.38\linewidth]{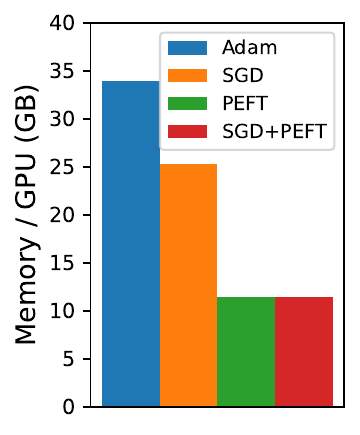}
    \\
    \includegraphics[width=0.75\linewidth]{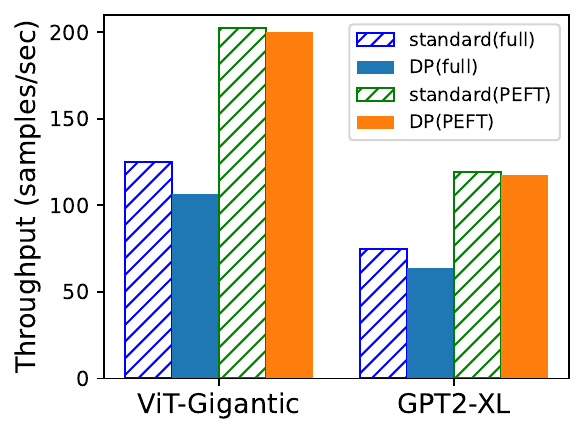}
    \caption{Efficiency of DP-ZeRO with lighter training. Upper: $>2\times$ speedup with lighter training; ViT-5B, $B=1$.
    % DP-ZeRO3
    Lower: DP-ZeRO benefits more than ZeRO from PEFT.
    % DP-ZeRO1, $B=4$.
    }
\vspace{-0.2cm}
    \label{fig:vit limits}
    \vspace{-1.2cm}
\end{wrapfigure}

DP-ZeRO can employ low-memory optimizers and train on fewer parameters, therefore vastly reducing the memory and communication cost. On a single node, we demonstrate that DP-ZeRO actually benefits (more than standard ZeRO and single-GPU training) from lighter training, following from our discussion in \Cref{sec:time efficiency}.

\paragraph{Low-memory optimizers}
Low-memory optimizers can boost the training efficiency at the cost of accuracy degradation. 
For example, SGD requires only 1/3 optimizer states of Adam and significantly saves the memory; 1-bit Adam \citep{tang20211} and signSGD compress the gradient and reduce the communication volume up to $32\times$.

\paragraph{Fewer trainable parameters}
In the fine-tuning phase, as we analyzed in \Cref{sec:PEFT}, PEFT improves both the local computation and the communication volume. Hence, DP-ZeRO allows PEFT on ViT and GPT to be $\approx 2\times$ faster than full fine-tuning, whereas the single GPU acceleration is $\leq 1.5\times$.

\begin{remark}
We leverage DP-ZeRO3 with SGD to train ViT-10B (full parameters) and ViT-22B (PEFT; 1M trainable parameters) on one node. See \Cref{app:settings} for configurations.
\end{remark}

\vspace{-0.3cm}
\subsection{Three stages of DP-ZeRO}
\vspace{-0.15cm}
DP-ZeRO supports all stages of ZeRO under different implementations including DeepSpeed (default) and FSDP.

\begin{figure}[!htb]
    \centering
    \includegraphics[width=0.49\linewidth]{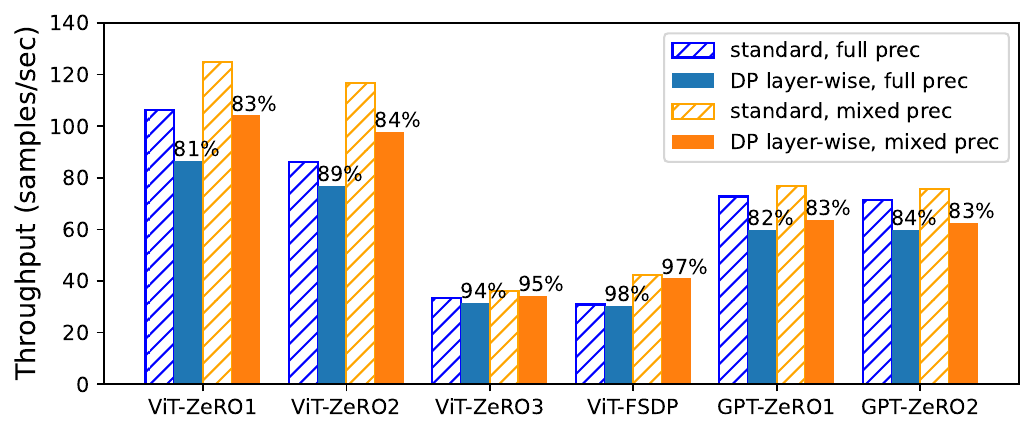}
\includegraphics[width=0.49\linewidth]{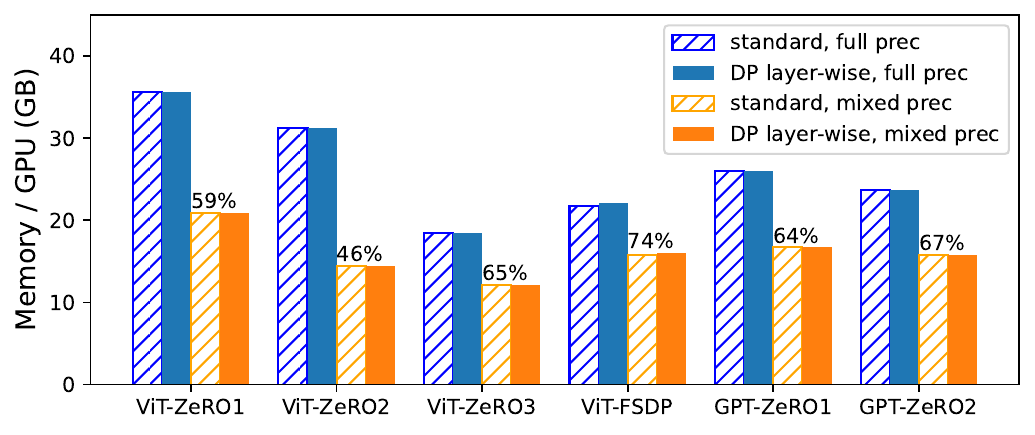}
\vspace{-0.25cm}
    \caption{Efficiency of DP-ZeRO
    %(1/2/3) 
    on ViT-Gigantic and GPT2-XL under different implementations.}
    \label{fig:multiple stages}
\vspace{-0.2cm}
\end{figure}

In \Cref{fig:multiple stages}, the efficiency of DP-ZeRO catches up with the standard ZeRO when we move up the stages. For instance of ViT-Gigantic, the throughput increases from 83\% by DP-ZeRO1 to 95-97\% by DP-ZeRO3. Following \eqref{eq:relative DP speed}, we can attribute the relatively fast training of DP-ZeRO to the increase cost of communication, especially in DP-ZeRO3. Additionally, we observe that the throughput of DP-ZeRO1/2 improves to over 95\% on 4 nodes, as predicted by \Cref{sec:number GPU}. Notice that 
% We highlight that GPT is fully supported by our DP-ZeRO for all three stages and implementations, and 
we save DP-ZeRO3 of GPT to \Cref{sec: scalability} on super-large scale.

\subsection{Scalability of DP-ZeRO}
\label{sec: scalability}
We evaluate the scalability of DP-ZeRO3 in terms of large sequence length (2048), large model size ($7\sim 100$B), and large number of GPUs (up to 256). We use A100 with 80GB memory, as well as the activation check-pointing and ModelP.

\begin{figure}[!htb]
    \centering
    \includegraphics[height=3.5cm]{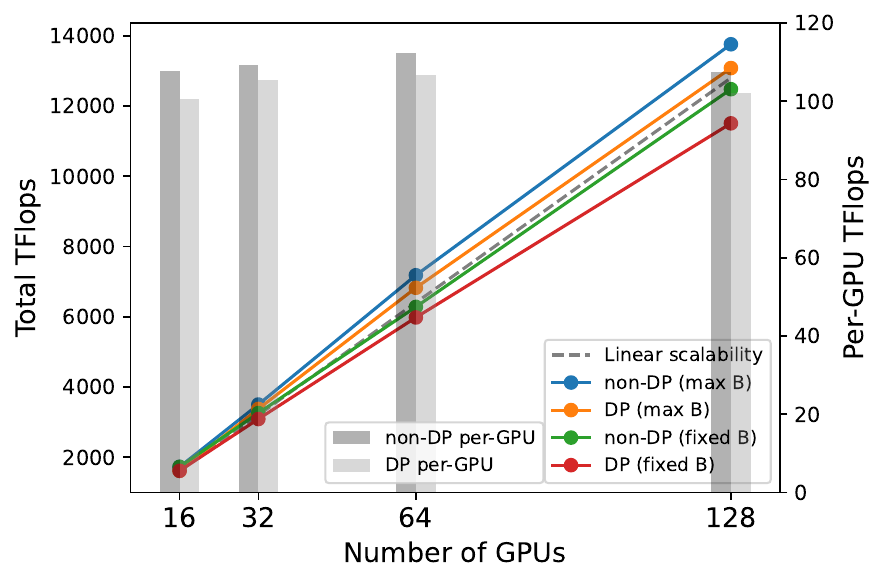}
    \includegraphics[height=3.5cm]{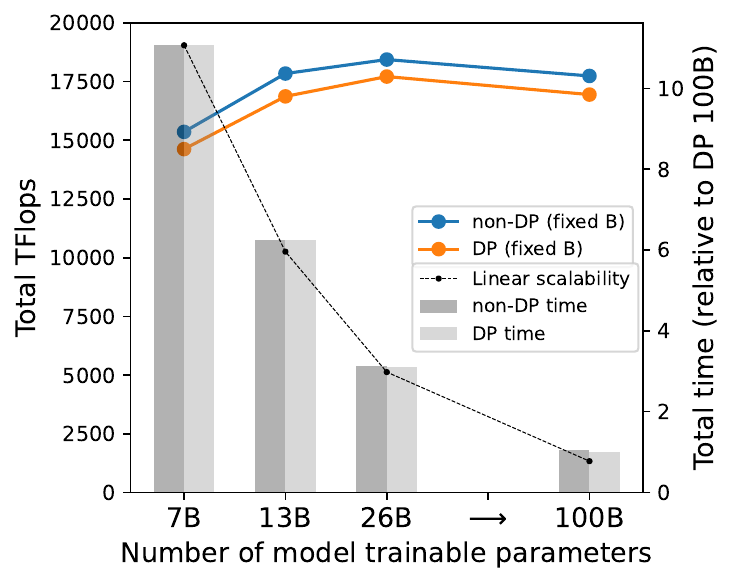}
    \vspace{-0.3cm}
    \caption{Scalability of DP-ZeRO3 on 26B model (left) and 256 GPUs (right). `max B' means we fit the maximum micro-batch in each GPU and `fixed B' means $B=2$.
    % The bars are per-GPU speed (max B).
    }
    \label{fig:DP increase GPUs}
\end{figure}

In \Cref{fig:DP increase GPUs} (left), we observe that for a fixed model with 26B trainable parameters, DP-ZeRO is super-linearly scalable to the number of GPUs, achieving $>95\%$ speed of the standard ZeRO. Here super-linearity is a property of ZeRO (see Figure 3 in \cite{rajbhandari2020zero}) which allows more GPUs to shard the model states (and reduce the per-GPU memory cost) more aggressively, and to train faster since the micro-batch size is larger. Furthermore, in \Cref{fig:DP increase GPUs} (right), for a fixed number of GPUs, DP-ZeRO is linearly scalable to the model size, achieving the same speed as the standard ZeRO. In short, DP-ZeRO is almost equal to ZeRO in terms of training efficiency in super-large scale.

% Fixed batch size, increase model size up to 7/13/26/52/175B, fix number of GPU (total FLOP, per-GPU FLOP). DP nonDP >>>>> we may want to separate the figure to also plot fixed b or non-DP per-GPU TFlops!!! 

\begin{remark}
In comparison to DP-ZeRO, DataP (DP or standard) at most fits 5B models \citep{rajbhandari2020zero} in 80GB memory, regardless of the number of GPUs. We cannot compare to DP-PipeP in \cite{anonymous2023exploring} because the codebase and experiment details (e.g. number of trainable parameters and sequence length) are not publicly available. Nevertheless, since DP-ZeRO resembles the efficiency of standard ZeRO, it suffices to demonstrate the usefulness of DP-ZeRO by comparing ZeRO to PipeP.
%{compare to DP pipeline, avoid param, avoid long sequence
\end{remark}

\section{Discussion}
In this work, we develope DP-ZeRO that enables the optimization of models up to $100$B trainable parameters, thus allowing DP distributed learning to be as efficient and scalable as the standard one. We believe this is a significant milestone to pave the path towards DP foundation models, especially for its open-source nature (link to be released).

We emphasize that, since DP only modifies the back-propagation, our DP-ZeRO is orthogonal to any large-scale training techniques that are not tied to back-propagation: for example, activation check-pointing, CPU offloading, weight/activation quantization \citep{dettmers2023qlora, xiao2023smoothquant}, tensor parallelism \citep{narayanan2021efficient} and other techniques yet to come. 
% The combination of DP-ZeRO with such techniques can serve as future direction for engineering. 
Note that to make fair comparisons in this paper, DP and non-DP optimization are implemented without fusing the operations such as tensor multiplications. Therefore, the efficiency of DP deep learning can be further and ultimately improved, from an engineering perspective, by implementing the DP back-propagation with operator fusion via CUDA kernels and C++ coding.
% This approach is different to the vanilla non-DP optimization, which fuses multiple operations into a single kernel to boost the training speed, at the cost of a heavier memory burden. It is desirable to close this gap between DP optimization and vanilla non-DP optimization, by implementing the DP back-propagation via CUDA kernels and C++ code.

\section{Acknowledgement} We thank Dhananjay Ram, Zhenghui Jin and Cole Hawkins for technical support on the AWS cluster, Xinwei Zhang for validating part of our code, Yu-Xiang Wang for the insightful initial discussion, and the hosts of FSDP (especially Min Xu) and DeepSpeed Github repositories for debugging part of our experiments.

\newpage
\bibliographystyle{iclr2024_conference}
\bibliography{references}

\begin{thebibliography}{58}
\providecommand{\natexlab}[1]{#1}
\providecommand{\url}[1]{\texttt{#1}}
\expandafter\ifx\csname urlstyle\endcsname\relax
  \providecommand{\doi}[1]{doi: #1}\else
  \providecommand{\doi}{doi: \begingroup \urlstyle{rm}\Url}\fi

\bibitem[Abadi et~al.(2016)Abadi, Chu, Goodfellow, McMahan, Mironov, Talwar,
  and Zhang]{abadi2016deep}
Martin Abadi, Andy Chu, Ian Goodfellow, H~Brendan McMahan, Ilya Mironov, Kunal
  Talwar, and Li~Zhang.
\newblock Deep learning with differential privacy.
\newblock In \emph{Proceedings of the 2016 ACM SIGSAC conference on computer
  and communications security}, pp.\  308--318, 2016.

\bibitem[Brown et~al.(2020)Brown, Mann, Ryder, Subbiah, Kaplan, Dhariwal,
  Neelakantan, Shyam, Sastry, Askell, et~al.]{brown2020language}
Tom Brown, Benjamin Mann, Nick Ryder, Melanie Subbiah, Jared~D Kaplan, Prafulla
  Dhariwal, Arvind Neelakantan, Pranav Shyam, Girish Sastry, Amanda Askell,
  et~al.
\newblock Language models are few-shot learners.
\newblock \emph{Advances in neural information processing systems},
  33:\penalty0 1877--1901, 2020.

\bibitem[Bu et~al.(2020)Bu, Dong, Long, and Su]{bu2020deep}
Zhiqi Bu, Jinshuo Dong, Qi~Long, and Weijie~J Su.
\newblock Deep learning with gaussian differential privacy.
\newblock \emph{Harvard data science review}, 2020\penalty0 (23), 2020.

\bibitem[Bu et~al.(2021)Bu, Gopi, Kulkarni, Lee, Shen, and
  Tantipongpipat]{bu2021fast}
Zhiqi Bu, Sivakanth Gopi, Janardhan Kulkarni, Yin~Tat Lee, Hanwen Shen, and
  Uthaipon Tantipongpipat.
\newblock Fast and memory efficient differentially private-sgd via jl
  projections.
\newblock \emph{Advances in Neural Information Processing Systems}, 34, 2021.

\bibitem[Bu et~al.(2022{\natexlab{a}})Bu, Mao, and Xu]{bu2022scalable}
Zhiqi Bu, Jialin Mao, and Shiyun Xu.
\newblock Scalable and efficient training of large convolutional neural
  networks with differential privacy.
\newblock \emph{Advances in Neural Information Processing Systems},
  35:\penalty0 38305--38318, 2022{\natexlab{a}}.

\bibitem[Bu et~al.(2022{\natexlab{b}})Bu, Wang, Zha, and
  Karypis]{bu2022dpbitfit}
Zhiqi Bu, Yu-Xiang Wang, Sheng Zha, and George Karypis.
\newblock Differentially private bias-term only fine-tuning of foundation
  models.
\newblock In \emph{Workshop on Trustworthy and Socially Responsible Machine
  Learning, NeurIPS 2022}, 2022{\natexlab{b}}.

\bibitem[Bu et~al.(2023{\natexlab{a}})Bu, Wang, Zha, and
  Karypis]{bu2022automatic}
Zhiqi Bu, Yu-Xiang Wang, Sheng Zha, and George Karypis.
\newblock Automatic clipping: Differentially private deep learning made easier
  and stronger.
\newblock \emph{Advances in Neural Information Processing Systems},
  2023{\natexlab{a}}.

\bibitem[Bu et~al.(2023{\natexlab{b}})Bu, Wang, Zha, and
  Karypis]{bu2022differentially}
Zhiqi Bu, Yu-Xiang Wang, Sheng Zha, and George Karypis.
\newblock Differentially private optimization on large model at small cost.
\newblock In \emph{International Conference on Machine Learning}, pp.\
  3192--3218. PMLR, 2023{\natexlab{b}}.

\bibitem[Chen et~al.(2016)Chen, Xu, Zhang, and Guestrin]{chen2016training}
Tianqi Chen, Bing Xu, Chiyuan Zhang, and Carlos Guestrin.
\newblock Training deep nets with sublinear memory cost.
\newblock \emph{arXiv preprint arXiv:1604.06174}, 2016.

\bibitem[De et~al.(2022)De, Berrada, Hayes, Smith, and Balle]{de2022unlocking}
Soham De, Leonard Berrada, Jamie Hayes, Samuel~L Smith, and Borja Balle.
\newblock Unlocking high-accuracy differentially private image classification
  through scale.
\newblock \emph{arXiv preprint arXiv:2204.13650}, 2022.

\bibitem[Dettmers et~al.(2023)Dettmers, Pagnoni, Holtzman, and
  Zettlemoyer]{dettmers2023qlora}
Tim Dettmers, Artidoro Pagnoni, Ari Holtzman, and Luke Zettlemoyer.
\newblock Qlora: Efficient finetuning of quantized llms.
\newblock \emph{arXiv preprint arXiv:2305.14314}, 2023.

\bibitem[Dong et~al.(2019)Dong, Roth, and Su]{dong2019gaussian}
Jinshuo Dong, Aaron Roth, and Weijie~J Su.
\newblock Gaussian differential privacy.
\newblock \emph{arXiv preprint arXiv:1905.02383}, 2019.

\bibitem[Dosovitskiy et~al.(2020)Dosovitskiy, Beyer, Kolesnikov, Weissenborn,
  Zhai, Unterthiner, Dehghani, Minderer, Heigold, Gelly,
  et~al.]{dosovitskiy2020image}
Alexey Dosovitskiy, Lucas Beyer, Alexander Kolesnikov, Dirk Weissenborn,
  Xiaohua Zhai, Thomas Unterthiner, Mostafa Dehghani, Matthias Minderer, Georg
  Heigold, Sylvain Gelly, et~al.
\newblock An image is worth 16x16 words: Transformers for image recognition at
  scale.
\newblock In \emph{International Conference on Learning Representations}, 2020.

\bibitem[Dwork et~al.(2006)Dwork, McSherry, Nissim, and
  Smith]{dwork2006calibrating}
Cynthia Dwork, Frank McSherry, Kobbi Nissim, and Adam Smith.
\newblock Calibrating noise to sensitivity in private data analysis.
\newblock In \emph{Theory of cryptography conference}, pp.\  265--284.
  Springer, 2006.

\bibitem[Foley \& Danskin(2017)Foley and Danskin]{foley2017ultra}
Denis Foley and John Danskin.
\newblock Ultra-performance pascal gpu and nvlink interconnect.
\newblock \emph{IEEE Micro}, 37\penalty0 (2):\penalty0 7--17, 2017.

\bibitem[Frostig et~al.(2018)Frostig, Johnson, and Leary]{frostig2018compiling}
Roy Frostig, Matthew~James Johnson, and Chris Leary.
\newblock Compiling machine learning programs via high-level tracing.
\newblock \emph{Systems for Machine Learning}, 4\penalty0 (9), 2018.

\bibitem[Goodfellow(2015)]{goodfellow2015efficient}
Ian Goodfellow.
\newblock Efficient per-example gradient computations.
\newblock \emph{arXiv preprint arXiv:1510.01799}, 2015.

\bibitem[Gopi et~al.(2021)Gopi, Lee, and Wutschitz]{gopi2021numerical}
Sivakanth Gopi, Yin~Tat Lee, and Lukas Wutschitz.
\newblock Numerical composition of differential privacy.
\newblock \emph{Advances in Neural Information Processing Systems}, 34, 2021.

\bibitem[He et~al.(2022)He, Li, Yu, Zhang, Kulkarni, Lee, Backurs, Yu, and
  Bian]{anonymous2023exploring}
Jiyan He, Xuechen Li, Da~Yu, Huishuai Zhang, Janardhan Kulkarni, Yin~Tat Lee,
  Arturs Backurs, Nenghai Yu, and Jiang Bian.
\newblock Exploring the limits of differentially private deep learning with
  group-wise clipping.
\newblock \emph{arXiv preprint arXiv:2212.01539}, 2022.

\bibitem[He et~al.(2016)He, Zhang, Ren, and Sun]{he2016deep}
Kaiming He, Xiangyu Zhang, Shaoqing Ren, and Jian Sun.
\newblock Deep residual learning for image recognition.
\newblock In \emph{Proceedings of the IEEE conference on computer vision and
  pattern recognition}, pp.\  770--778, 2016.

\bibitem[Houlsby et~al.(2019)Houlsby, Giurgiu, Jastrzebski, Morrone,
  De~Laroussilhe, Gesmundo, Attariyan, and Gelly]{houlsby2019parameter}
Neil Houlsby, Andrei Giurgiu, Stanislaw Jastrzebski, Bruna Morrone, Quentin
  De~Laroussilhe, Andrea Gesmundo, Mona Attariyan, and Sylvain Gelly.
\newblock Parameter-efficient transfer learning for nlp.
\newblock In \emph{International Conference on Machine Learning}, pp.\
  2790--2799. PMLR, 2019.

\bibitem[Hu et~al.(2021)Hu, Shen, Wallis, Allen-Zhu, Li, Wang, Wang, and
  Chen]{hu2021lora}
Edward~J Hu, Yelong Shen, Phillip Wallis, Zeyuan Allen-Zhu, Yuanzhi Li, Shean
  Wang, Lu~Wang, and Weizhu Chen.
\newblock Lora: Low-rank adaptation of large language models.
\newblock \emph{arXiv preprint arXiv:2106.09685}, 2021.

\bibitem[Huang et~al.(2019)Huang, Cheng, Bapna, Firat, Chen, Chen, Lee, Ngiam,
  Le, Wu, et~al.]{huang2019gpipe}
Yanping Huang, Youlong Cheng, Ankur Bapna, Orhan Firat, Dehao Chen, Mia Chen,
  HyoukJoong Lee, Jiquan Ngiam, Quoc~V Le, Yonghui Wu, et~al.
\newblock Gpipe: Efficient training of giant neural networks using pipeline
  parallelism.
\newblock \emph{Advances in neural information processing systems}, 32, 2019.

\bibitem[Ishii et~al.(2018)Ishii, Foley, Anderson, Dally, Dearth, Dennison,
  Hummel, and Schafer]{ishii2018nvswitch}
Alex Ishii, Denis Foley, E~Anderson, B~Dally, G~Dearth, L~Dennison, M~Hummel,
  and J~Schafer.
\newblock Nvswitch and dgx-2 nvlink-switching chip and scale-up compute server.
\newblock In \emph{Hot Chips}, 2018.

\bibitem[Katharopoulos et~al.(2020)Katharopoulos, Vyas, Pappas, and
  Fleuret]{katharopoulos2020transformers}
Angelos Katharopoulos, Apoorv Vyas, Nikolaos Pappas, and Fran{\c{c}}ois
  Fleuret.
\newblock Transformers are rnns: Fast autoregressive transformers with linear
  attention.
\newblock In \emph{International conference on machine learning}, pp.\
  5156--5165. PMLR, 2020.

\bibitem[Kingma \& Ba(2014)Kingma and Ba]{kingma2014adam}
Diederik~P Kingma and Jimmy Ba.
\newblock Adam: A method for stochastic optimization.
\newblock \emph{arXiv preprint arXiv:1412.6980}, 2014.

\bibitem[Koskela et~al.(2020)Koskela, J{\"a}lk{\"o}, and
  Honkela]{koskela2020computing}
Antti Koskela, Joonas J{\"a}lk{\"o}, and Antti Honkela.
\newblock Computing tight differential privacy guarantees using fft.
\newblock In \emph{International Conference on Artificial Intelligence and
  Statistics}, pp.\  2560--2569. PMLR, 2020.

\bibitem[Kurakin et~al.(2022)Kurakin, Chien, Song, Geambasu, Terzis, and
  Thakurta]{kurakin2022toward}
Alexey Kurakin, Steve Chien, Shuang Song, Roxana Geambasu, Andreas Terzis, and
  Abhradeep Thakurta.
\newblock Toward training at imagenet scale with differential privacy.
\newblock \emph{arXiv preprint arXiv:2201.12328}, 2022.

\bibitem[Lee \& Kifer(2021)Lee and Kifer]{lee2020scaling}
Jaewoo Lee and Daniel Kifer.
\newblock Scaling up differentially private deep learning with fast per-example
  gradient clipping.
\newblock \emph{Proceedings on Privacy Enhancing Technologies}, 2021\penalty0
  (1), 2021.

\bibitem[Li et~al.(2019)Li, Song, Chen, Li, Liu, Tallent, and
  Barker]{li2019evaluating}
Ang Li, Shuaiwen~Leon Song, Jieyang Chen, Jiajia Li, Xu~Liu, Nathan~R Tallent,
  and Kevin~J Barker.
\newblock Evaluating modern gpu interconnect: Pcie, nvlink, nv-sli, nvswitch
  and gpudirect.
\newblock \emph{IEEE Transactions on Parallel and Distributed Systems},
  31\penalty0 (1):\penalty0 94--110, 2019.

\bibitem[Li et~al.()Li, Zhao, Varma, Salpekar, Noordhuis, Li, Paszke, Smith,
  Vaughan, Damania, et~al.]{li13pytorch}
Shen Li, Yanli Zhao, Rohan Varma, Omkar Salpekar, Pieter Noordhuis, Teng Li,
  Adam Paszke, Jeff Smith, Brian Vaughan, Pritam Damania, et~al.
\newblock Pytorch distributed: Experiences on accelerating data parallel
  training.
\newblock \emph{Proceedings of the VLDB Endowment}, 13\penalty0 (12).

\bibitem[Li et~al.(2021)Li, Tramer, Liang, and Hashimoto]{li2021large}
Xuechen Li, Florian Tramer, Percy Liang, and Tatsunori Hashimoto.
\newblock Large language models can be strong differentially private learners.
\newblock \emph{arXiv preprint arXiv:2110.05679}, 2021.

\bibitem[McMahan et~al.(2018)McMahan, Ramage, Talwar, and
  Zhang]{mcmahan2018learning}
H~Brendan McMahan, Daniel Ramage, Kunal Talwar, and Li~Zhang.
\newblock Learning differentially private recurrent language models.
\newblock In \emph{International Conference on Learning Representations}, 2018.

\bibitem[Mehta et~al.(2022)Mehta, Thakurta, Kurakin, and
  Cutkosky]{mehta2022large}
Harsh Mehta, Abhradeep Thakurta, Alexey Kurakin, and Ashok Cutkosky.
\newblock Large scale transfer learning for differentially private image
  classification.
\newblock \emph{arXiv preprint arXiv:2205.02973}, 2022.

\bibitem[Micikevicius et~al.(2018)Micikevicius, Narang, Alben, Diamos, Elsen,
  Garcia, Ginsburg, Houston, Kuchaiev, Venkatesh,
  et~al.]{micikevicius2018mixed}
Paulius Micikevicius, Sharan Narang, Jonah Alben, Gregory Diamos, Erich Elsen,
  David Garcia, Boris Ginsburg, Michael Houston, Oleksii Kuchaiev, Ganesh
  Venkatesh, et~al.
\newblock Mixed precision training.
\newblock In \emph{International Conference on Learning Representations}, 2018.

\bibitem[Narayanan et~al.(2021)Narayanan, Shoeybi, Casper, LeGresley, Patwary,
  Korthikanti, Vainbrand, Kashinkunti, Bernauer, Catanzaro,
  et~al.]{narayanan2021efficient}
Deepak Narayanan, Mohammad Shoeybi, Jared Casper, Patrick LeGresley, Mostofa
  Patwary, Vijay Korthikanti, Dmitri Vainbrand, Prethvi Kashinkunti, Julie
  Bernauer, Bryan Catanzaro, et~al.
\newblock Efficient large-scale language model training on gpu clusters using
  megatron-lm.
\newblock In \emph{Proceedings of the International Conference for High
  Performance Computing, Networking, Storage and Analysis}, pp.\  1--15, 2021.

\bibitem[Nvidia()]{NVIDIA}
Nvidia.
\newblock The most powerful end-to-end ai and hpc data center platform.
\newblock
  \url{https://www.nvidia.com/en-us/data-center/tensor-cores/?ref=blog.paperspace.com}.

\bibitem[Radford et~al.(2019)Radford, Wu, Child, Luan, Amodei, Sutskever,
  et~al.]{radford2019language}
Alec Radford, Jeffrey Wu, Rewon Child, David Luan, Dario Amodei, Ilya
  Sutskever, et~al.
\newblock Language models are unsupervised multitask learners.
\newblock \emph{OpenAI blog}, 1\penalty0 (8):\penalty0 9, 2019.

\bibitem[Rajbhandari et~al.(2020)Rajbhandari, Rasley, Ruwase, and
  He]{rajbhandari2020zero}
Samyam Rajbhandari, Jeff Rasley, Olatunji Ruwase, and Yuxiong He.
\newblock Zero: Memory optimizations toward training trillion parameter models.
\newblock In \emph{SC20: International Conference for High Performance
  Computing, Networking, Storage and Analysis}, pp.\  1--16. IEEE, 2020.

\bibitem[Rajbhandari et~al.(2021)Rajbhandari, Ruwase, Rasley, Smith, and
  He]{rajbhandari2021zero}
Samyam Rajbhandari, Olatunji Ruwase, Jeff Rasley, Shaden Smith, and Yuxiong He.
\newblock Zero-infinity: Breaking the gpu memory wall for extreme scale deep
  learning.
\newblock In \emph{Proceedings of the International Conference for High
  Performance Computing, Networking, Storage and Analysis}, pp.\  1--14, 2021.

\bibitem[Ren et~al.(2021)Ren, Rajbhandari, Aminabadi, Ruwase, Yang, Zhang, Li,
  and He]{ren2021zero}
Jie Ren, Samyam Rajbhandari, Reza~Yazdani Aminabadi, Olatunji Ruwase, Shuangyan
  Yang, Minjia Zhang, Dong Li, and Yuxiong He.
\newblock $\{$ZeRO-Offload$\}$: Democratizing $\{$Billion-Scale$\}$ model
  training.
\newblock In \emph{2021 USENIX Annual Technical Conference (USENIX ATC 21)},
  pp.\  551--564, 2021.

\bibitem[Shen et~al.(2021)Shen, Zhang, Zhao, Yi, and Li]{shen2021efficient}
Zhuoran Shen, Mingyuan Zhang, Haiyu Zhao, Shuai Yi, and Hongsheng Li.
\newblock Efficient attention: Attention with linear complexities.
\newblock In \emph{Proceedings of the IEEE/CVF winter conference on
  applications of computer vision}, pp.\  3531--3539, 2021.

\bibitem[Subramani et~al.(2021)Subramani, Vadivelu, and
  Kamath]{subramani2021enabling}
Pranav Subramani, Nicholas Vadivelu, and Gautam Kamath.
\newblock Enabling fast differentially private sgd via just-in-time compilation
  and vectorization.
\newblock \emph{Advances in Neural Information Processing Systems}, 34, 2021.

\bibitem[Tang et~al.(2021)Tang, Gan, Awan, Rajbhandari, Li, Lian, Liu, Zhang,
  and He]{tang20211}
Hanlin Tang, Shaoduo Gan, Ammar~Ahmad Awan, Samyam Rajbhandari, Conglong Li,
  Xiangru Lian, Ji~Liu, Ce~Zhang, and Yuxiong He.
\newblock 1-bit adam: Communication efficient large-scale training with
  adam’s convergence speed.
\newblock In \emph{International Conference on Machine Learning}, pp.\
  10118--10129. PMLR, 2021.

\bibitem[Touvron et~al.(2023{\natexlab{a}})Touvron, Lavril, Izacard, Martinet,
  Lachaux, Lacroix, Rozi{\`e}re, Goyal, Hambro, Azhar,
  et~al.]{touvron2022llama}
Hugo Touvron, Thibaut Lavril, Gautier Izacard, Xavier Martinet, Marie-Anne
  Lachaux, Timoth{\'e}e Lacroix, Baptiste Rozi{\`e}re, Naman Goyal, Eric
  Hambro, Faisal Azhar, et~al.
\newblock Llama: Open and efficient foundation language models.
\newblock \emph{arXiv preprint arXiv:2302.13971}, 2023{\natexlab{a}}.

\bibitem[Touvron et~al.(2023{\natexlab{b}})Touvron, Martin, Stone, Albert,
  Almahairi, Babaei, Bashlykov, Batra, Bhargava, Bhosale,
  et~al.]{touvron2023llama}
Hugo Touvron, Louis Martin, Kevin Stone, Peter Albert, Amjad Almahairi, Yasmine
  Babaei, Nikolay Bashlykov, Soumya Batra, Prajjwal Bhargava, Shruti Bhosale,
  et~al.
\newblock Llama 2: Open foundation and fine-tuned chat models.
\newblock \emph{arXiv preprint arXiv:2307.09288}, 2023{\natexlab{b}}.

\bibitem[Vaswani et~al.(2017)Vaswani, Shazeer, Parmar, Uszkoreit, Jones, Gomez,
  Kaiser, and Polosukhin]{vaswani2017attention}
Ashish Vaswani, Noam Shazeer, Niki Parmar, Jakob Uszkoreit, Llion Jones,
  Aidan~N Gomez, {\L}ukasz Kaiser, and Illia Polosukhin.
\newblock Attention is all you need.
\newblock \emph{Advances in neural information processing systems}, 30, 2017.

\bibitem[Wang et~al.(2020)Wang, Li, Khabsa, Fang, and Ma]{wang2020linformer}
Sinong Wang, Belinda~Z Li, Madian Khabsa, Han Fang, and Hao Ma.
\newblock Linformer: Self-attention with linear complexity.
\newblock \emph{arXiv preprint arXiv:2006.04768}, 2020.

\bibitem[Xiao et~al.(2023)Xiao, Lin, Seznec, Wu, Demouth, and
  Han]{xiao2023smoothquant}
Guangxuan Xiao, Ji~Lin, Mickael Seznec, Hao Wu, Julien Demouth, and Song Han.
\newblock {S}mooth{Q}uant: Accurate and efficient post-training quantization
  for large language models.
\newblock In \emph{Proceedings of the 40th International Conference on Machine
  Learning}, 2023.

\bibitem[Xie et~al.(2018)Xie, Lin, Wang, Wang, and Zhou]{xie2018differentially}
Liyang Xie, Kaixiang Lin, Shu Wang, Fei Wang, and Jiayu Zhou.
\newblock Differentially private generative adversarial network.
\newblock \emph{arXiv preprint arXiv:1802.06739}, 2018.

\bibitem[Yang et~al.(2022)Yang, Zhang, Chen, and Liu]{yang2022normalized}
Xiaodong Yang, Huishuai Zhang, Wei Chen, and Tie-Yan Liu.
\newblock Normalized/clipped sgd with perturbation for differentially private
  non-convex optimization.
\newblock \emph{arXiv preprint arXiv:2206.13033}, 2022.

\bibitem[Yousefpour et~al.(2021)Yousefpour, Shilov, Sablayrolles, Testuggine,
  Prasad, Malek, Nguyen, Ghosh, Bharadwaj, Zhao, Cormode, and Mironov]{opacus}
Ashkan Yousefpour, Igor Shilov, Alexandre Sablayrolles, Davide Testuggine,
  Karthik Prasad, Mani Malek, John Nguyen, Sayan Ghosh, Akash Bharadwaj,
  Jessica Zhao, Graham Cormode, and Ilya Mironov.
\newblock Opacus: {U}ser-friendly differential privacy library in {PyTorch}.
\newblock \emph{arXiv preprint arXiv:2109.12298}, 2021.

\bibitem[Yu et~al.(2021)Yu, Naik, Backurs, Gopi, Inan, Kamath, Kulkarni, Lee,
  Manoel, Wutschitz, et~al.]{yu2021differentially}
Da~Yu, Saurabh Naik, Arturs Backurs, Sivakanth Gopi, Huseyin~A Inan, Gautam
  Kamath, Janardhan Kulkarni, Yin~Tat Lee, Andre Manoel, Lukas Wutschitz,
  et~al.
\newblock Differentially private fine-tuning of language models.
\newblock \emph{arXiv preprint arXiv:2110.06500}, 2021.

\bibitem[Zaken et~al.(2022)Zaken, Goldberg, and Ravfogel]{zaken2022bitfit}
Elad~Ben Zaken, Yoav Goldberg, and Shauli Ravfogel.
\newblock Bitfit: Simple parameter-efficient fine-tuning for transformer-based
  masked language-models.
\newblock In \emph{Proceedings of the 60th Annual Meeting of the Association
  for Computational Linguistics (Volume 2: Short Papers)}, pp.\  1--9, 2022.

\bibitem[Zhai et~al.(2022)Zhai, Kolesnikov, Houlsby, and
  Beyer]{zhai2022scaling}
Xiaohua Zhai, Alexander Kolesnikov, Neil Houlsby, and Lucas Beyer.
\newblock Scaling vision transformers.
\newblock In \emph{Proceedings of the IEEE/CVF Conference on Computer Vision
  and Pattern Recognition}, pp.\  12104--12113, 2022.

\bibitem[Zhang et~al.(2022)Zhang, Zheng, Wang, Chiu, Karypis, Chilimbi, Li, and
  Jin]{zhang2022mics}
Zhen Zhang, Shuai Zheng, Yida Wang, Justin Chiu, George Karypis, Trishul
  Chilimbi, Mu~Li, and Xin Jin.
\newblock Mics: near-linear scaling for training gigantic model on public
  cloud.
\newblock \emph{Proceedings of the VLDB Endowment}, 16\penalty0 (1):\penalty0
  37--50, 2022.

\bibitem[Zhao et~al.()Zhao, Varma, Huang, Li, Xu, and Desmaison]{FSDP}
Yanli Zhao, Rohan Varma, Chien-Chin Huang, Shen Li, Min Xu, and Alban
  Desmaison.
\newblock Introducing pytorch fully sharded data parallel (fsdp) api.
\newblock
  \url{https://pytorch.org/blog/introducing-pytorch-fully-sharded-data-parallel-api/}.

\bibitem[Zhu et~al.(2022)Zhu, Dong, and Wang]{zhu2021optimal}
Yuqing Zhu, Jinshuo Dong, and Yu-Xiang Wang.
\newblock Optimal accounting of differential privacy via characteristic
  function.
\newblock In \emph{International Conference on Artificial Intelligence and
  Statistics}, pp.\  4782--4817. PMLR, 2022.

\end{thebibliography}

\clearpage
\appendix
\section{The Book-Keeping (BK) algorithm}
\label{app:BK}

\subsection{Efficient computation of per-sample gradient norms}
\label{app:mixed ghost norm}
The mixed ghost norm \cite{bu2022scalable} is the state-of-the-art technique to compute the per-sample gradient norm of the \textbf{weight}, almost for free. It hybridizes two basic techniques -- the per-sample gradient instantiation and the ghost norm -- to compute the Frobenius norm of weight gradient, 
\begin{align}
\left\|\a_{l,i}^\top\frac{\partial L}{\partial\s_{l,i}}\right\|^2\overset{\text{per-sample grad}}{=\joinrel=\joinrel=\joinrel=}\|\frac{\partial L_i}{\partial \W_{l}}\|_\text{Fro}^2\overset{\text{ghost norm}}{=\joinrel=\joinrel=\joinrel=}\text{vec}\left(\frac{\partial L}{\partial \s_{l,i}}\frac{\partial L}{\partial \s_{l,i}}^\top\right)\cdot\text{vec}(\a_{l,i} \a_{l,i}^\top)
\label{eq:mixed ghost norm}
\end{align}
where "vec" flattens the tensor to an one-dimensional vector. In words, \Cref{eq:mixed ghost norm} gives two equations that are equivalent mathematically, but significantly different in efficiency:
\begin{itemize}
\item $\|X\|_\text{Fro}^2=\left\|A^\top B\right\|^2$ firstly computes $A^\top B$ and then its norm.
\item $\|X\|_\text{Fro}^2=\text{vec}\left(AA^\top\right)\cdot\text{vec}(BB^\top)$ firstly computes $AA^\top,BB^\top\in\R^{TT}$ and then their dot product.
\end{itemize}
% That is, the second technique computes $X$'s norm without ever instantiating the variable $X$, similar to the idea of Gaussian elimination that solves $A^{-1}b$ without inverting $A$.

% The complexity analysis \cite{bu2022scalable,bu2022differentially} has shown that $\|X\|_\text{Fro}^2=\left\|A^\top B\right\|^2$ requires $pd$ space complexity and  $2Tpd$ time complexity, while $\|X\|_\text{Fro}^2=\text{vec}\left(AA^\top\right)\cdot\text{vec}(BB^\top)$ requires $2T^2$ space complexity and $2T^2(p + d)$ time complexity.

% To put this into perspective, to compute the per-sample gradient norm of the weight, we denote $X=\frac{\partial L_i}{\partial \W_l}, A=\a_{l,i}, B=\frac{\partial L}{\partial \s_{l,i}}$. Consequently, the two techniques are :

In summary, the mixed ghost norm always applies the cheaper of two techniques at each layer of a neural network.

Finally, we note that the per-sample gradient norm of the \textbf{bias} is computed differently. This is because
$$\frac{\partial L_i}{\partial b_l}=\mathbf{1}^\top \frac{\partial L}{\partial \s_{l,i}}$$
is not actually a product of tensors like $X=A^\top B$. In fact, the multiplication with $\mathbf{1}$ turns out to be a summation along the first dimension, and it suffices to use per-sample gradient instantiation for the bias.

\subsection{Book-keeping the output gradient}
The BK algorithm uses two rounds of back-propagation (though each round only takes half the complexity, hence the total complexity of DP back-propagation matches the non-DP back-propagation). Therefore, output gradients $\frac{\partial L}{\partial \s_{l,i}}$ are kept to avoid repeated computation. Notice that the output gradient are relatively cheap to book-keep (see Figure 4 and Figure 5 in \cite{bu2022differentially}).

\section{Component-wise time complexity of DP-ZeRO}
\label{app:time complexity}

In this section, we explain the time complexity of each part in \Cref{tab:complexity full finetune}, and demonstrate how the complexity can be different under different settings.

\textbf{Forward propagation: }The matrix multiplication during forward propagation results in $2BT\Psi_\text{model}$ complexity (see \cite{bu2022scalable}). Notice that, if the activation check-pointing is used, essentially two rounds of forward propagation take place in one iteration. Hence the time complexity doubles and becomes $4BT\Psi_\text{model}$.

\textbf{Back-propagation:} This contains two sub-processes: the output gradients are computed at all layers, taking $2BT\Psi_\text{model}$ complexity; the parameter gradients are computed only at trainable layers (a few if doing PEFT), taking $2BT\Psi_\text{train}$ complexity. Clearly, in full parameter training, the total is $4BT\Psi_\text{model}$, and in PEFT, about $2BT\Psi_\text{model}$.

\textbf{Attention: }The time complexity of attention is $O(BT^2)$ in \cite{vaswani2017attention}, where $T$ is the sequence length (a.k.a. token length). When $T$ is large, e.g. training with long context like $T=8192$, this cost is prohibitively high. In this regard, a line of researches have proposed linear complexity attention, including but not limited to \cite{wang2020linformer, katharopoulos2020transformers,shen2021efficient}.

\textbf{Communication: }For algorithms that don't shard the model, such as data parallelism and ZeRO1/2, the communication is only used to send gradients and optimizer states. Hence the communication volume is proportional to the number of trainable parameters $O(\Psi_\text{train})$. Otherwise, for algorithms such as ZeRO3 and tensor parallelism, the communication volume is proportional to the number of total parameters $O(\Psi_\text{model})$, because the forward propagation needs to gather the parameters from many GPUs. This makes a big difference in PEFT when $\Psi_\text{train}\ll \Psi_\text{model}$.

\section{Loss scaling in mixed-precision training}
\label{app:loss scaling}
We write the per-sample gradient with loss scaling $S$ as
$$\frac{\partial C_i L_i}{\partial \W_l}=C_i\frac{1}{S}\cdot\left(\a_{l,i}^\top \big(S\cdot\frac{\partial L}{\partial \s_{l,i}}\big)\right)$$
This covers the standard gradient ($C_i=1$) and DP gradient (e.g. $C_i=1/||\g_i||$, computed by the mixed ghost norm in \Cref{app:BK}), in which $S$ enlarges the output gradient to avoid underflow, and $\frac{1}{S}$ shrinks the parameter gradient to the correct magnitude.

Recall that a standard mixed-precision training (with loss scaling) uses steps $1\to 2\to 3\to 5\to 6$ \footnote{See \url{https://docs.nvidia.com/deeplearning/performance/mixed-precision-training/index.html\#lossscaling}.}, or $1\to 3\to 6$ without loss scaling.

\setstcolor{green}
\begin{enumerate}
    \item Forward propagation (fp16 weights and activations) and get the loss.
    \item \st{Scaling up the loss by a factor $S$.}
    \item Backward propagation on the scaled loss (fp16 parameters and their gradients).
    \item {\color{blue} Per-sample gradient clipping (sensitivity $=1$) and noising for DP}.
    % \item \st{Scheduling the scale: if no Inf or NAN in parameter gradients, increase $S$; else, reduce $S$.}
    \item \st{Scaling down the parameter gradient by a factor of $1/S$.}
    \item Update the parameters with their gradients.
\end{enumerate}
\setstcolor{black}

If we follow the same procedure under the DP regime, say using a hook function to be called after back-propagation creates the gradients like in Opacus \citep{opacus}, Private-Transformers \citep{li2021large}, FastDP \citep{bu2022differentially}, then the per-sample clipping factor is scaled up $S$ times so as to normalize the gradient. Hence per-sample gradient clipping has already played the role of scaling down. If we scale down the gradient for a second time, the gradient is incorrectly over-shrunk. This is the case in \cite{yu2021differentially} and in the alternative implementation of \cite[Appendix T]{li2021large} (see also \Cref{fig:DP loss scaling}). To be sure, this approach is still DP, but the performance does not match fp32 DP training correctly, and usually degrades too much to be useful.

One walk-around is to prevent per-sample gradient clipping to scale down the gradients and let step 5 do its job, i.e. $1\to 2\to 3\to 4^*\to 5\to 6$.  We note that \cite[Appendix T]{li2021large} follows this path (though no experiment results or codes are available at the time of writing) by modifying step 4: clipping threshold (sensitivity)$=S$ instead of $1$, so that the clipped gradient is $S$ times larger than the DP f32 training, to be scaled down by step 5. However, this introduces additional design decisions and does not prevent overflow when using fp16 (due to step 2, see \Cref{tab:loss scale}).

Another walk-around is to delete step 5 and let per-sample gradient clipping do its job, i.e. $1\to 2\to 3\to 4\to 6$. However, this approach is harder to implement because in the standard process step 2 and 5 are simultaneously enabled or disabled. Also we cannot prevent overflow when using fp16 as we still use step 2.

Therefore, we propose to not use loss scaling (or equivalently we set $S=1$ statically for all steps) during DP mixed-precision training, i.e. $1\to 3\to 4\to 6$. Although, by not using step 2, we cannot prevent underflow when using fp16, this is much less a problem compared to overflow: underflow (treating small values as 0) makes the training less accurate but does not fail the training like overflow (treating large values as NAN). Lastly, the underflow issue is perfectly mitigated by bf16, which we recommend for DP mixed-precision training whenever possible . 

\begin{table}[!htb]
    \centering
    \begin{tabular}{c|c|c|c|c}
    &steps&fp16 issue&note&reference\\\hline
    standard & 136 & underflow&&\cite{micikevicius2018mixed} \\
    standard & 12356& none &&\cite{micikevicius2018mixed}
\\
    DP & 123456& overflow & incorrect due to over-shrinking &\cite{li2021large}
\\
    DP & $1234^*56$& overflow & different clipping threshold &\cite{li2021large}
\\
    DP & 1346& underflow & perfect with bf16&ours
\\
    DP & 12346& overflow & hard to implement&ours
    \end{tabular}
    \caption{Mixed-precision training with DP or not.}
\end{table}

\section{Experiment settings}
\label{app:settings}
Datasets: To evaluate the efficiency, it suffices to declare the data's dimension (e.g. micro-batch size and feature dimension) without specifying the dataset (though sometimes specifying the dataset means declaring the dimension, e.g. MNIST usually means 28*28 pixels). This is the norm in system papers such as \cite{rajbhandari2020zero,rajbhandari2021zero,FSDP}. In this work, vision models are trained with 224*224 pixels at ImageNet scale; GPT models are trained with sequence length 100, except in \Cref{fig:DP increase GPUs} where sequence length is 2048.

\Cref{fig:DP loss scaling} and \Cref{tab:loss scale}: We train ViT-large (300M parameters) and CIFAR100, 5 epochs, learning rate 5e-4, logical batch size 1000.

\Cref{fig:vit limits}: To fit as large a model as possible, we set $B=1$ and use SGD. We set 48 attention heads, 21 layers, MLP=4*width (also known as embedding dimension), and modify width for all models. For instance, ViT-10B uses width=$768*22$, ViT-22B uses width=$768*34$.

\Cref{fig:DP increase GPUs}: We train AdamW with layer-wise clipping. DP distributed learning is based on MiCS (ZeRO3) using bf16 mixed-precision training. %shard8; for GPU in 2/4/8/16 nodes, max b=2,3,4,4. 
Most of GPT configuration is the same as \cite{touvron2022llama} (Table 2) in terms of embedding dimension, attention heads and number of layers. However, GPT-100B uses the configuration from \cite{brown2020language} (Table 2.1) but a smaller width.

\iffalse
\section{DP-ZeRO compatible with per-sample gradient clipping styles}

Consider 4 layers in ().

non-DP backward:
\begin{align*}
&\text{all-gather (1)}\to \text{output grad (1)}\to\text{param grad (1)}\to\text{reduce-scatter (1)}
\\
\to&\text{all-gather (2)}\to \text{output grad (2)}\to\text{param grad (2)}\to\text{reduce-scatter (2)}
\\
\to&\text{all-gather (3)}\to \text{output grad (3)}\to\text{param grad (3)}\to\text{reduce-scatter (3)}
\\
\to&\text{all-gather (4)}\to \text{output grad (4)}\to\text{param grad (4)}\to\text{reduce-scatter (4)}
\end{align*}
The activations have be deleted, but can be re-computed from checkpoint. We have each layer's clip norm, and can compute all-layer clipping factor.

DP backward:
\begin{align*}
&\text{all-gather (1)}\to \text{output grad (1, keep)}\to\text{all-gather (2)}\to \text{output grad (2, keep)}
\\\to&\text{all-gather (3)}\to \text{output grad (3, keep)}\to\text{all-gather (4)}\to \text{output grad (4, keep)}
\\
\to&\text{clip factor (whole)}
\\
\to&\text{param grad (1)}\to\text{reduce-scatter (1)}\to\text{param grad (2)}\to\text{reduce-scatter (2)}
\\\to&\text{param grad (3)}\to\text{reduce-scatter (3)}\to\text{param grad (4)}\to\text{reduce-scatter (4)}
\end{align*}
\fi

\section{Codebase design}
\subsection{With forward \& backward hooks}
Hooks\footnote{See \url{https://pytorch.org/tutorials/beginner/former\_torchies/nnft\_tutorial.html\#forward-and-backward-function-hooks}.} are important functions to enrich the deep learning optimization. To be specific, there are
\begin{enumerate}
    \item forward modular hook (nn.register\_forward\_hook), 
    % triggered during forward propagation after activations have been computed.
    \item backward modular hook (nn.register\_backward\_hook), 
    % triggered during backward propagation after gradients have been computed.
    \item backward tensor hook (tensor.register\_hook). 
    % triggered during backward propagation after gradients have been computed.
\end{enumerate}

DP libraries including Opacus \cite{opacus}, Private-transformers \cite{li2021large}, Private-Vision \cite{bu2022scalable}, FastDP \cite{bu2022differentially,bu2022automatic}, FastGradClip \cite{lee2020scaling} and so on, use modular hooks to modify the standard optimization. However, ZeRO libraries including DeepSpeed and FSDP use tensor hooks. This difference in the types of hooks and many other differences (e.g. both ZeRO libraries and DP libraries modify the optimizer's step function) cause non-trivial problems when combining DP with ZeRO. For example, to keep DP optimization as efficient as the standard, it is necessary to not waste time on computing the non-private gradient. However, if we skip such computation, then ZeRO's tensor hook will not be triggered and the corresponding distributed-learning-related operations cannot carry on. For another example, because DP and ZeRO add different types of hooks, the number of hooks is larger than either optimization and they slows down the training: consider an 100-layer network, each layer with weight and bias (2 tensors), then DP-ZeRO in this subsection needs 100 modular hooks and 200 tensor hooks, adding to a total of 300 hooks. In addition, the Book-Keeping algorithm (in FastDP) in its original form cannot be implemented together with ZeRO3, because all model states are partitioned including the output gradients which are meant to be book-kept. To work around this requires rewriting the distributed solution's communication mechanism, and if successful, still requiring additional communication cost during the second back-propagation. Similar problems are present for Opacus and FastGradClip, which instantiates per-sample gradients that will be partitioned in ZeRO2/3 and requires additional communication cost when gathered to create the privatized gradient.

As a consequence, the hooks are fully supported on DP-ZeRO1 and partially supported on DP-ZeRO2/3 under the layer-wise clipping.

\subsection{Without hooks}
Instead of registering hooks on top of the original (non-DP) back-propagation, we can directly modify the back-propagation following \Cref{app:BK}: e.g., given the activation and output gradient,
$$
\frac{\partial C_i L_i}{\partial \W_l}=\a_{(l),i}^\top\frac{\partial L}{\partial \s_{(l),i}}\Big/\sqrt{\text{vec}\left(\frac{\partial L}{\partial \s_{(l),i}}\frac{\partial L}{\partial \s_{(l),i}}^\top\right)\cdot\text{vec}(\a_{(l),i} \a_{(l),i}^\top)}
$$
This approach requires rewriting the back-propagation for each layer type (linear, embedding, convolution, normalization, ...) and can be done at different levels (Pytorch, C++, CUDA kernel).

\subsection{User interface}
DP-ZeRO can be enabled by one piece of code: after the model is instantiated,
\begin{verbatim}
privacy_engine = PrivacyEngine(model, 
                batch_size=256, sample_size=50000, 
                epochs=3, target_epsilon=3)
\end{verbatim}
The codebase is designed not to modify the optimizer, hence DP-ZeRO can work with arbitrary optimizer. Because of this design, our DP-ZeRO will not distinguish micro-batches. This is different from the gradient accumulation in Opacus (version ==0.x) and Private-Vision, where only the last micro-batch is processed by "optimizer.step()" but all other micro-batches are processed by "optimizer.virtual\_step()". In other words, the noise $\sigma_\text{DP}N(0,I)$ is added on the last micro-batch, after the micro-batches are accumulated. But DP-ZeRO adds the noise on each micro-batch equally. Note that the noise level per micro-batch is $\sigma_\text{DP}/\sqrt{N_d}$ if a random seed is set across $N_d$ GPUs, or $\sigma_\text{DP}/N_d$ otherwise.

% \subsection{Noise addition}

\end{document}